\documentclass[10pt,twocolumn]{article}

\usepackage[letterpaper,top=0.72in,bottom=0.78in,left=0.68in,right=0.68in,columnsep=0.26in]{geometry}
\usepackage[T1]{fontenc}
\usepackage[utf8]{inputenc}
\usepackage{lmodern}
\usepackage{microtype}
\usepackage{amsmath}
\usepackage{amssymb}
\usepackage{graphicx}
\usepackage{booktabs}
\usepackage{multirow}
\usepackage[table]{xcolor}
\usepackage[round,authoryear]{natbib}
\usepackage{caption}
\usepackage{url}
\usepackage{hyperref}
\usepackage{titling}
\usepackage{titlesec}

\microtypesetup{protrusion=true,expansion=true}
\titleformat{\section}
  {\large\bfseries}
  {\thesection}
  {0.7em}
  {}
\titleformat{\subsection}
  {\normalsize\bfseries}
  {\thesubsection}
  {0.65em}
  {}
\titleformat{\subsubsection}
  {\normalsize\bfseries}
  {\thesubsubsection}
  {0.6em}
  {}
\titlespacing*{\section}
  {0pt}
  {1.4ex plus 0.3ex minus 0.2ex}
  {0.65ex plus 0.15ex}
\titlespacing*{\subsection}
  {0pt}
  {1.1ex plus 0.2ex minus 0.15ex}
  {0.4ex plus 0.1ex}
\titlespacing*{\subsubsection}
  {0pt}
  {0.9ex plus 0.2ex minus 0.1ex}
  {0.3ex}

\definecolor{gainGreen}{RGB}{32,140,82}
\definecolor{dropRed}{RGB}{190,65,75}
\definecolor{bestBlue}{RGB}{228,239,250}
\definecolor{secondRed}{RGB}{252,228,232}
\definecolor{groupGray}{RGB}{245,245,245}

\newcommand{\gain}[1]{%
  \rlap{\hspace{0.08em}%
  {\scriptsize\textcolor{gainGreen}{$\uparrow #1$}}}%
}

\newcommand{\dropval}[1]{%
  \rlap{\hspace{0.08em}%
  {\scriptsize\textcolor{dropRed}{$\downarrow #1$}}}%
}

\newcommand{\topcell}[1]{%
  \cellcolor{bestBlue}\textbf{#1}%
}

\newcommand{\secondcell}[1]{%
  \cellcolor{secondRed}\textbf{#1}%
}

\hypersetup{
  colorlinks=true,
  linkcolor=blue!50!black,
  citecolor=blue!50!black,
  urlcolor=blue!50!black,
  pdftitle={Look Ahead Before You Distill: Future Trajectory Validation of Teacher Guidance for Agentic On-Policy Distillation},
  pdfauthor={Chishui Chen, Yaoyou Fan, Te Sun, Yi Yang, Chenghao Sun, Delin Mao, Hongbo Qiao, Zuowei Zhang, Junxi Wang, Chenxing Sun, Yangen Hu, Lu Pan, Xuyang Liu, Linfeng Zhang}
}

\newcommand{\correspondingmark}{\ensuremath{\dagger}}

\makeatletter
\newcommand{\authornote}[1]{%
  \begingroup
  \renewcommand{\thefootnote}{}%
  \renewcommand{\footnoterule}{}%
  \long\def\@makefntext##1{\noindent ##1}%
  \footnotetext{#1}%
  \endgroup
}
\makeatother

\title{\Large\bfseries Look Ahead Before You Distill: Future Trajectory Validation of Teacher Guidance for Agentic On-Policy Distillation}
\author{
\normalsize\bfseries
Chishui Chen\textsuperscript{1,3,*} \quad
Yaoyou Fan\textsuperscript{1,4,*} \quad
Te Sun\textsuperscript{2,*} \quad
Yi Yang\textsuperscript{1,5,*} \quad
Chenghao Sun\textsuperscript{6}
\\
\normalsize\bfseries
Delin Mao\textsuperscript{2} \quad
Hongbo Qiao\textsuperscript{7} \quad
Zuowei Zhang\textsuperscript{8} \quad
Junxi Wang\textsuperscript{3} \quad
Chenxing Sun\textsuperscript{1,\correspondingmark}
\\
\normalsize\bfseries
Yangen Hu\textsuperscript{1} \quad
Lu Pan\textsuperscript{1} \quad
Xuyang Liu\textsuperscript{9} \quad
Linfeng Zhang\textsuperscript{2,\correspondingmark}
\\[0.60em]
\fontsize{9.5}{11}\selectfont\bfseries
\textsuperscript{1}Meituan LongCat Interaction \quad
\textsuperscript{2}Shanghai Jiao Tong University \quad
\textsuperscript{3}Fudan University
\\
\fontsize{9.5}{11}\selectfont\bfseries
\textsuperscript{4}Peking University \quad
\textsuperscript{5}Nanjing University \quad
\textsuperscript{6}University of Chinese Academy of Sciences
\\
\fontsize{9.5}{11}\selectfont\bfseries
\textsuperscript{7}Jilin University \quad
\textsuperscript{8}University of Science and Technology of China
\\
\fontsize{9.5}{11}\selectfont\bfseries
\textsuperscript{9}The Hong Kong Polytechnic University
}
\posttitle{\par\end{center}\vskip 0.3em}
\postauthor{\end{tabular}\par\end{center}\vskip 0.5em}
\date{}

\begin{document}
\maketitle

\authornote{%
\footnotesize\raggedright
\textsuperscript{*}Equal contribution. \quad
\textsuperscript{\correspondingmark}Corresponding authors.\par
Email: \href{mailto:chishui.chen@outlook.com}{chishui.chen@outlook.com}.\par
Work done during an internship at Meituan.\vspace{0.10in}%
}
\setcounter{footnote}{0}

\begin{abstract}
On-policy distillation (OPD) provides teacher supervision on states visited by the student, reducing the distribution gap between training and inference. However, in multi-turn agentic tasks, student deviations may accumulate over time, gradually moving the trajectory away from states where teacher guidance remains effective.
Our quantitative analysis further shows that high-disagreement states offer promising opportunities for teacher guidance, but determining whether such guidance is beneficial requires examining its effect on subsequent student trajectories. We propose \textbf{FutureBridge-OPD} (\textbf{FTB}), which executes a short teacher bridge at a high disagreement state and uses the resulting student continuation to assess whether the bridge increases the density of positive distillation signals relative to the teacher. On ALFWorld, WebShop, and ScienceWorld, under the main Qwen3-32B teacher to Qwen3-1.7B student setting, FTB outperforms vanilla OPD and TCOD by an average of \textbf{16.6} and \textbf{7.6} points, respectively, and remains effective across student scales and teacher settings. Our code is publicly available at \url{https://github.com/ChenChiShui/FutureBridge-OPD}.
\end{abstract}

\section{Introduction}
\label{sec:introduction}

On-policy distillation (OPD) has recently emerged as an effective approach for transferring the capabilities of large language models to smaller models~\cite{agarwal2024onpolicy}.
By providing teacher supervision on states visited by the student itself, OPD alleviates the distribution mismatch between offline distillation and actual inference~\cite{AReductionofImitation,ScheduledSampling}.
However, in multi-turn agentic tasks, early deviations made by the student alter the subsequent state distribution and accumulate along the trajectory, gradually weakening the effectiveness of later teacher supervision~\cite{sod,reopd}.

Existing agentic OPD methods mainly tackle this issue in two ways.
Methods such as TCOD and Guided-OPD regulate student takeover through teacher prefixes, curriculum schedules, or adaptive rollout depths~\cite{tcod,guided-opd,turnopd}.
Another line of work filters or reweights distillation signals using KL divergence, model confidence, or environment feedback~\cite{TIP,SCOPE,SAGE-OPD}.
The former reshapes the student-visited state distribution, whereas the latter estimates supervision value from local information at the current state.
However, both largely overlook one question:
\textbf{\textit{Can teacher guidance steer the student's future trajectory toward regions with higher teacher preference?}}

\begin{figure*}[t]
    \centering
    \includegraphics[width=\textwidth]{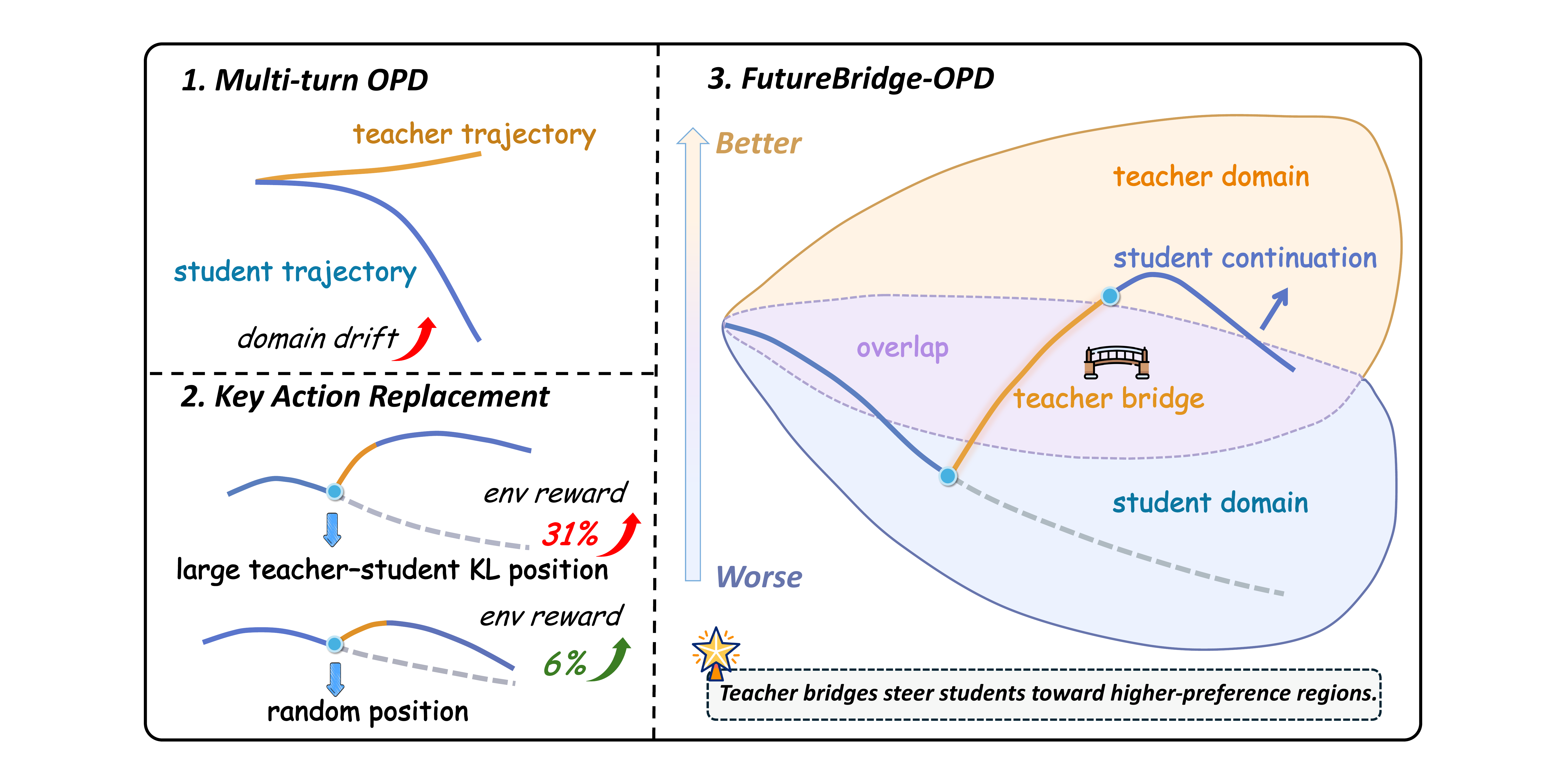}
    \caption{
    Motivation and overview of FutureBridge-OPD.
    (1) Student deviations accumulate over multi-turn interactions.
    (2) High KL positions offer promising intervention opportunities.
    (3) The induced student continuation validates the teacher bridge, which redirects the student toward teacher preferred regions.
    }
    \label{fig:intro}
\end{figure*}

To study this question, we analyze 1,000 failed trajectories from ALFWorld, WebShop, and ScienceWorld~\cite{alfworld,webshop,scienceworld}.
We replay the history prefix, replace the student action at positions with large teacher--student disagreement, and then let the same student policy complete the remaining interaction. Further details are given in Appendix~\ref{sec:appendix_motivation}.
This analysis reveals two useful properties of teacher guidance in multi-turn agentic OPD:

\noindent \textbf{(I) Key Action Replacement Improves Outcomes.}
As illustrated in the second panel of Figure~\ref{fig:intro}, replacing the student action with the teacher action at positions with large teacher--student KL divergence improves environment rewards in approximately \textbf{31\%} of the trajectories, compared with only \textbf{6\%} when actions are replaced at random positions.
However, even among these high-KL positions, approximately \textbf{19\%} of the interventions still degrade trajectory performance.
This indicates that teacher--student KL divergence is useful for identifying candidate intervention positions, but insufficient for determining whether the corresponding teacher guidance is beneficial.

\noindent \textbf{(II) Teacher Guidance Redirects Students Toward Teacher-Preferred Regions.}
We further examine how teacher guidance affects the continuation autonomously generated by the student.
After adopting the teacher action, the proportion of teacher preferred tokens in the subsequent student trajectory increases by approximately \textbf{10.8\%} in relative terms, conceptually illustrated in the third panel of Figure~\ref{fig:intro} as the student continuation shifting toward the teacher preferred region.
This result suggests that the effect of appropriate local guidance is not confined to the current decision, but persists across subsequent interactions.
Therefore, assessing the value of candidate teacher guidance also requires looking ahead to the future student trajectory that it induces.

We thus propose \textbf{FutureBridge-OPD} (\textbf{FTB}), a two-stage framework that locates candidate interventions through teacher--student disagreement and validates teacher guidance through the induced future student trajectory.
FTB first identifies a candidate position and executes a \emph{teacher bridge} by replacing the student decision with a teacher action.
The same student policy then generates a short continuation from the bridged state, and the resulting future trajectory determines whether the guidance redirects subsequent student behavior toward regions with higher teacher preference and should be retained for distillation.

Our main contributions are summarized as follows:
\begin{itemize}
    \setlength{\topsep}{0.35em}
    \setlength{\itemsep}{0.35em}
    \setlength{\parsep}{0pt}
    \setlength{\partopsep}{0pt}
    \setlength{\parskip}{0pt}
    \item \textbf{Systematic Analysis of Teacher Guidance in Agentic OPD.}
    We analyze the interplay among student rollout depth, teacher--student disagreement, and future distillation signals in multi-turn agentic OPD. We show that high-disagreement interventions more often improve outcomes and steer future student trajectories toward denser positive distillation signals.

    \item \textbf{A Framework for Future Validation of Teacher Guidance.}
    We propose FutureBridge-OPD, which executes a teacher bridge at a candidate decision and validates it through the induced student continuation before distillation.

    \item \textbf{Consistent Gains across Tasks and Models.}
    Across ALFWorld, WebShop, and ScienceWorld, FutureBridge-OPD improves the average success rate by \textbf{16.6} and \textbf{7.6} points over vanilla OPD~\cite{agarwal2024onpolicy} and TCOD-B2F~\cite{tcod}, respectively, and remains effective across student scales and teacher settings.
\end{itemize}

\section{Related Work}

\noindent\textbf{On-Policy Distillation for Agents.} Knowledge distillation transfers the predictive distribution of a large
teacher to a smaller student, enabling model compression and capability
transfer~\cite{distillingtheknowledge}. Distillation at the sequence
level extends this principle to autoregressive sequence
generation~\cite{Sequence-Level-knowledge}, while MiniLLM further
investigates reverse KL objectives tailored to large language model
generation~\cite{minillm}. On-policy distillation (OPD) provides
teacher supervision on data generated and states visited by the student,
thereby reducing the distribution mismatch between offline distillation
and actual inference~\cite{agarwal2024onpolicy}.

Recent studies further examine the mechanisms, applicable conditions,
and potential failure modes of OPD under long horizon generation
~\cite{rethinkopd,RevisitingOn-PolicyDistillation,pruneopd,yourteacher}.
For multi-turn agent tasks, TCOD uses a temporal curriculum to control
how deeply the student takes over the trajectory~\cite{tcod}, while
Guided-OPD mixes teacher and student turns according to a curriculum
schedule~\cite{guided-opd}. Other studies improve trajectory
construction and teacher guidance in multi-turn OPD through prefix
replay, short-to-long rollouts, adaptive rollout depths, adjustment at
the turn level, or environment feedback
~\cite{reopd,shortopd,turnopd,ATOD,SAGE-OPD}.

\noindent\textbf{Selective Supervision and Reweighting in OPD.} Another line of work focuses on selecting or reweighting distillation
supervision across different tokens, turns, or reasoning steps.
Existing methods use signals such as generation position, teacher
entropy, student uncertainty, causal lookahead, teacher--student
divergence, trajectory outcomes, or model perplexity to identify
high value supervision or adjust its distillation strength
~\cite{prefixopd,eopd,egrsd,TIP,SCOPE,sod,tropd}. These methods generally estimate supervision value primarily from the current response, local model distributions, or outcomes of trajectories already generated by the student itself.

\noindent\textbf{Future Information for Supervision and Credit Assignment.} In language model reasoning and policy optimization, several studies
use future continuations to estimate the downstream effects of local
decisions. Existing approaches reshape advantages at the token level using
future KL divergence, perform detailed credit assignment through
counterfactual perturbations or gradient approximations, resample
continuations from intermediate reasoning states, or combine local
uncertainty with future continuation signals for branching and guidance
~\cite{FIPO,oar,rrpo,APPO,topd}. These studies demonstrate the value
of future information for evaluating local decisions.

\section{Method}

\subsection{Preliminary: On-Policy Distillation}

We consider a finite horizon, multi-turn interaction between an agent
and an environment. At turn $t$, the agent receives an observation
$o_t$, generates a response $a_t$, and executes it as a single
environment action. The environment then returns the next observation
$o_{t+1}$. We denote the interaction history by
$h_t=(o_1,a_1,\ldots,o_{t-1},a_{t-1},o_t)$ and a complete trajectory
by $\tau=(o_1,a_1,\ldots,o_T,a_T,o_{T+1})$.

On-policy distillation (OPD) aligns a student policy $\pi_\theta$
with a teacher policy $\pi_\phi$ on states visited by the student,
thereby reducing the mismatch between training and inference as follows:
\begin{equation}
\mathcal{L}_{\mathrm{OPD}}(\theta)
=
\mathbb{E}_{\tau\sim p_{\pi_\theta}}
\left[
\sum_{t=1}^{T}
D_{\mathrm{KL}}
\left(
\pi_\theta(\cdot\mid h_t)
\,\|\,
\pi_\phi(\cdot\mid h_t)
\right)
\right].
\label{eq:opd}
\end{equation}

For multi-turn tasks, we adopt the Back-to-Forward curriculum
(B2F) from TCOD. Given a successful trajectory of length $L$, B2F
replays its first $L-k$ turns and lets the student complete the
remaining interaction, with $k$ gradually increasing during
training. Following TCOD's public implementation, trajectories are
collected using a frozen student policy $\pi_{\bar\theta}$, and
Eq.~\eqref{eq:opd} is optimized with a sampled token estimator during training.

For a student response
$a_t=(x_{t,1},\ldots,x_{t,M_t})$, let
$c_{t,i}=(h_t,x_{t,<i})$ be the context of token $x_{t,i}$. We denote
the raw sampled distillation advantage by
\begin{equation}
A_{t,i}
=
\beta
\left[
\log\pi_\phi(x_{t,i}\mid c_{t,i})
-
\log\pi_{\bar\theta}(x_{t,i}\mid c_{t,i})
\right],
\label{eq:sampled_advantage}
\end{equation}
where $\beta>0$ is the distillation coefficient. A positive value
indicates that the teacher assigns the sampled token higher likelihood
than the frozen student. We explicitly call $A_{t,i}$ \emph{raw} here
to distinguish it from clipped or normalized variants used during
optimization.

\subsection{FutureBridge-OPD}

\begin{figure*}[t]
    \centering
    \includegraphics[width=\textwidth]{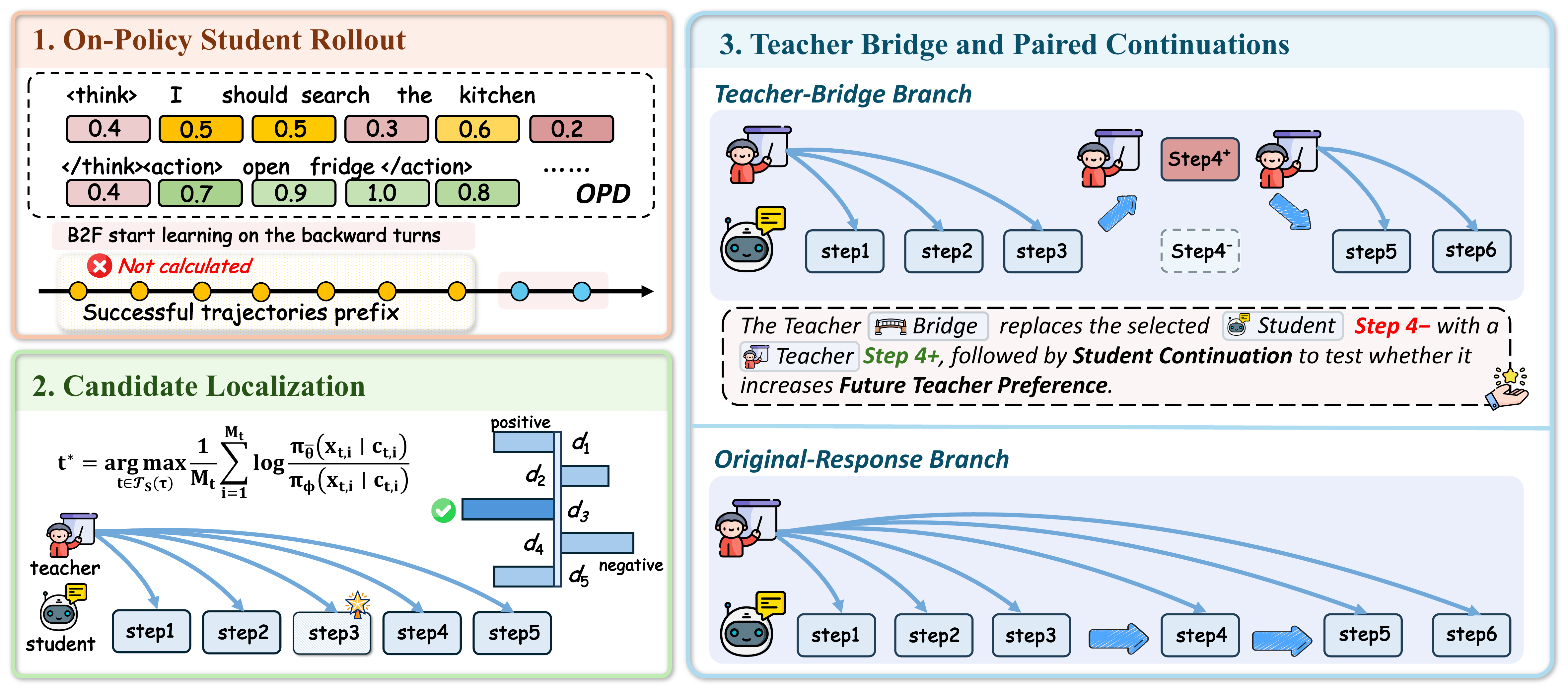}
    \caption{
    Overview of FutureBridge-OPD:
    (1) on-policy student rollouts provide distillation signals;
    (2) maximum teacher--student disagreement identifies a candidate bridge turn;
    (3) paired continuations validate the teacher bridge.
    }
    \label{fig:pipeline}
\end{figure*}

FTB uses a local sampled signal to propose a candidate teacher
intervention and paired future continuations to validate its
downstream effect, as illustrated in Figure~\ref{fig:pipeline}.

\noindent\textbf{Candidate Localization.}
Given a B2F student rollout $\tau$, we score the response actually
executed at each turn controlled by the student:
\begin{equation}
\widehat d_t
=
\frac{1}{M_t}
\sum_{i=1}^{M_t}
\log
\frac{
\pi_{\bar\theta}(x_{t,i}\mid c_{t,i})
}{
\pi_\phi(x_{t,i}\mid c_{t,i})
}.
\label{eq:realized_score}
\end{equation}
Let
$\overline A_t=M_t^{-1}\sum_{i=1}^{M_t}A_{t,i}$.
By autoregressive factorization,
\[
\widehat d_t
=
\frac{1}{M_t}
\log
\frac{\pi_{\bar\theta}(a_t\mid h_t)}
     {\pi_\phi(a_t\mid h_t)}
=
-\frac{1}{\beta}\overline A_t.
\]
We therefore select
\begin{equation}
t^\star
=
\arg\max_{t\in\mathcal T_S(\tau)}
\widehat d_t
=
\arg\min_{t\in\mathcal T_S(\tau)}
\overline A_t,
\label{eq:bridge_turn}
\end{equation}
where $\mathcal T_S(\tau)$ contains the turns generated by the student after
the replayed teacher prefix, excluding the final turn.

Thus, FTB selects the realized student response with the lowest
average distillation advantage relative to the teacher. Here, $\widehat d_t$ directly evaluates the response
that induced the observed environment transition. It is a directional
sampled score and is used only to localize a potentially correctable
turn; whether the teacher replacement is beneficial is determined by
the future validation below. At most one bridge is attempted per trajectory, bounding validation cost and isolating a single intervention.

\noindent\textbf{Teacher Bridge and Paired Continuations.}
Let $a_{t^\star}^{\mathrm{stu}}$ denote the original student response.
Conditioned on the same history $h_{t^\star}$, the teacher generates
a bridge response $a_{t^\star}^{\mathrm{br}}$ using the student's
decoding configuration. We restore the environment state at
$t^\star$, execute $a_{t^\star}^{\mathrm{br}}$ in place of
$a_{t^\star}^{\mathrm{stu}}$, and return control to the frozen
student $\pi_{\bar\theta}$ for an $H$-turn continuation
$\xi^{\mathrm{br}}$. The corresponding continuation following the
original student response is denoted by $\xi^{\mathrm{base}}$.

Both continuations start from the same history before intervention and,
after the alternative actions, are generated by the same frozen
student. Their difference therefore reflects the downstream effect
of replacing the realized student response with the teacher bridge.

\noindent\textbf{Future Validation Gate.}
For any sequence generated by the student
$y=(y_1,\ldots,y_{|y|})$, we define
\begin{equation}
\rho(y)
=
\frac{1}{|y|}
\sum_{i=1}^{|y|}
\mathbf{1}\!\left[A_i(y)>0\right],
\label{eq:teacher_preference}
\end{equation}
where $A_i(y)$ denotes Eq.~\eqref{eq:sampled_advantage} evaluated on
token $y_i$ and its context. Equivalently, $\rho(y)$ is the fraction
of tokens to which the teacher assigns higher likelihood than the
frozen student. It measures the teacher preferred token ratio along the generated student continuation. This ratio is normalized by the total number
of generated tokens, facilitating comparisons between continuations
of different lengths and preventing a small number of large
log probability differences from dominating the aggregate statistic.

The bridge acceptance indicator is
\begin{equation}
g(\tau)
=
\mathbf{1}
\left[
\rho(\xi^{\mathrm{br}})
>
\rho(\xi^{\mathrm{base}})
\right].
\label{eq:bridge_gate}
\end{equation}
A bridge is retained only when its induced student continuation has a
higher teacher preferred token ratio than the corresponding original continuation.

The paired continuations are used only as validation evidence. For
an accepted bridge, only $a_{t^\star}^{\mathrm{br}}$ is added to the
B2F experience buffer. Beyond the successful reference trajectories
inherited from B2F, this procedure uses no additional environment
reward or task success label.

\subsection{Future Teacher Preference}
\label{sec:interpretation}

Let $\xi_H(h,a;\pi_{\bar\theta})$ denote the continuation obtained by
executing action $a$ under history $h$ and subsequently rolling out
the frozen student for $H$ turns. We define an auxiliary
future teacher preference value by
\begin{equation}
U^H_{\phi,\bar\theta}(h,a)
=
\mathbb{E}
\left[
\rho\left(\xi_H(h,a;\pi_{\bar\theta})\right)
\right].
\end{equation}
It measures the expected teacher preferred token ratio in the student
continuation induced by $a$.

For the teacher bridge and the original student response, define the future teacher preference gain
\begin{equation}
\Delta^H_{\phi,\bar\theta}
=
U^H_{\phi,\bar\theta}(h_{t^\star},a^{\mathrm{br}})
-
U^H_{\phi,\bar\theta}(h_{t^\star},a^{\mathrm{stu}}),
\end{equation}
with the paired sample estimate
\begin{equation}
\widehat{\Delta}^H_{\phi,\bar\theta}
=
\rho(\xi^{\mathrm{br}})
-
\rho(\xi^{\mathrm{base}}).
\end{equation}
The gate retains bridges satisfying
$\widehat{\Delta}^H_{\phi,\bar\theta}>0$, i.e., bridges with a
positive sampled future teacher preference gain.

Candidate localization and future validation use complementary
aggregations of the same sampled OPD signal. The signed average
advantage localizes a realized response receiving weak
teacher relative support, while the teacher preferred token ratio characterizes
the student continuation induced by replacing that response.

Accepted teacher bridges are optimized with
\begin{equation}
\mathcal{L}_{\mathrm{bridge}}(\theta)
=
-
\mathbb{E}
\left[
\frac{g(\tau)}{M_{\mathrm{br}}}
\sum_{i=1}^{M_{\mathrm{br}}}
\operatorname{sg}[A_i^{\mathrm{br}}]
\log\pi_\theta(x_i^{\mathrm{br}}\mid c_i^{\mathrm{br}})
\right].
\end{equation}
At $\theta=\bar\theta$, treating the accepted bridge data and gate as
fixed, its gradient matches that of a sampled squared
teacher--student log probability discrepancy objective. The formal statement and proof are provided in Appendix~\ref{app:formal}.

\section{Experiments}
\subsection{Experimental Setup}

\begin{table*}[!t]
\centering

\small
\setlength{\tabcolsep}{5.5pt}

\begin{tabular*}{\textwidth}{
    @{\extracolsep{\fill}}
    l
    c
    cc
    cc
    @{\hspace{1.8em}}
}
\toprule

\multirow{2}{*}{\textbf{Method}}
&
\multicolumn{1}{c}{\textbf{ALFWorld}}
&
\multicolumn{2}{c}{\textbf{WebShop}}
&
\multicolumn{2}{c}{\textbf{ScienceWorld}}
\\

\cmidrule(lr){2-2}
\cmidrule(lr){3-4}
\cmidrule(lr){5-6}

&
\textbf{SR $\uparrow$}
&
\textbf{Score $\uparrow$}
&
\textbf{SR $\uparrow$}
&
\textbf{Score $\uparrow$}
&
\textbf{SR $\uparrow$}
\\

\midrule


\rowcolor{groupGray}
\multicolumn{6}{c}{
\textnormal{Qwen3-32B teacher $\rightarrow$ Qwen3-1.7B student}
}
\\

Student (zero-shot)
& 7.5
& 39.0
& 31.0
& 6.4
& 2.0
\\

Teacher (zero-shot)
& 50.0
& 52.8
& 51.0
& 28.8
& 12.0
\\

\addlinespace[1pt]

OPD
& 21.5 $\pm$ 3.0
& 32.6 $\pm$ 2.2
& 31.7 $\pm$ 2.9
& 22.2 $\pm$ 1.3
& 2.0 $\pm$ 1.5
\\

TCOD-F2B
& 30.7 $\pm$ 3.2\gain{9.2}
& 39.6 $\pm$ 1.7\gain{7.0}
& 38.3 $\pm$ 1.2\gain{6.6}
& \secondcell{26.1 $\pm$ 0.0}\gain{3.9}
& 5.8 $\pm$ 0.8\gain{3.8}
\\

TCOD-B2F
& \secondcell{35.0 $\pm$ 3.3}\gain{13.5}
& 41.8 $\pm$ 3.7\gain{9.2}
& 40.3 $\pm$ 3.3\gain{8.6}
& 22.3 $\pm$ 0.4\gain{0.1}
& \secondcell{6.7 $\pm$ 0.6}\gain{4.7}
\\

Guided-OPD
& 33.1 $\pm$ 5.4\gain{11.6}
& 44.6 $\pm$ 11.7\gain{12.0}
& 41.7 $\pm$ 10.6\gain{10.0}
& 23.5 $\pm$ 2.3\gain{1.3}
& 4.7 $\pm$ 0.6\gain{2.7}
\\

FTB w/o Bridge Exec.
& 31.4 $\pm$ 3.2\gain{9.9}
& \topcell{58.4 $\pm$ 3.4}\gain{25.8}
& \secondcell{55.3 $\pm$ 3.7}\gain{23.6}
& 18.7 $\pm$ 3.3\dropval{3.5}
& 3.7 $\pm$ 1.2\gain{1.7}
\\

\textbf{FTB (Ours)}
& \topcell{40.4 $\pm$ 3.4}\gain{18.9}
& \secondcell{58.1 $\pm$ 1.5}\gain{25.5}
& \topcell{57.7 $\pm$ 1.7}\gain{26.0}
& \topcell{26.4 $\pm$ 0.4}\gain{4.2}
& \topcell{6.8 $\pm$ 0.6}\gain{4.8}
\\

\midrule


\rowcolor{groupGray}
\multicolumn{6}{c}{
\textnormal{Qwen3-32B teacher $\rightarrow$ Qwen3-4B student}
}
\\

Student (zero-shot)
& 20.9
& 32.9
& 29.0
& 23.1
& 8.5
\\

Teacher (zero-shot)
& 50.0
& 52.8
& 51.0
& 28.8
& 12.0
\\

\addlinespace[1pt]

OPD
& 42.3 $\pm$ 4.9
& 43.2 $\pm$ 1.7
& 40.3 $\pm$ 2.6
& \topcell{29.0 $\pm$ 0.3}
& 7.3 $\pm$ 1.0
\\

TCOD-F2B
& 47.5 $\pm$ 4.6\gain{5.2}
& 49.7 $\pm$ 7.9\gain{6.5}
& 46.7 $\pm$ 6.8\gain{6.4}
& 23.9 $\pm$ 0.4\dropval{5.1}
& 11.3 $\pm$ 2.5\gain{4.0}
\\

TCOD-B2F
& 39.8 $\pm$ 4.8\dropval{2.5}
& \topcell{50.9 $\pm$ 1.2}\gain{7.7}
& 47.3 $\pm$ 2.6\gain{7.0}
& 24.5 $\pm$ 2.1\dropval{4.5}
& 9.5 $\pm$ 0.4\gain{2.2}
\\

Guided-OPD
& 47.0 $\pm$ 9.3\gain{4.7}
& 47.7 $\pm$ 5.8\gain{4.5}
& 47.0 $\pm$ 5.9\gain{6.7}
& 28.3 $\pm$ 1.7\dropval{0.7}
& 11.3 $\pm$ 2.0\gain{4.0}
\\

FTB w/o Bridge Exec.
& \topcell{48.5 $\pm$ 9.2}\gain{6.2}
& \secondcell{50.7 $\pm$ 0.3}\gain{7.5}
& \secondcell{48.3 $\pm$ 1.2}\gain{8.0}
& 24.6 $\pm$ 3.7\dropval{4.4}
& \secondcell{12.5 $\pm$ 3.3}\gain{5.2}
\\

\textbf{FTB (Ours)}
& \secondcell{48.3 $\pm$ 1.9}\gain{6.0}
& \topcell{50.9 $\pm$ 5.9}\gain{7.7}
& \topcell{50.0 $\pm$ 5.7}\gain{9.7}
& \secondcell{28.4 $\pm$ 2.9}\dropval{0.6}
& \topcell{14.0 $\pm$ 0.4}\gain{6.7}
\\

\bottomrule
\end{tabular*}

\caption{Main results with Qwen3-32B as the teacher, reported as mean $\pm$ standard deviation over three seeds; annotations show absolute changes over OPD, while \textbf{blue} and \textbf{red} backgrounds denote the best and second-best trained methods, respectively.}

\label{tab:main_results}

\end{table*}

\noindent\textbf{Benchmarks and Models.}
We conduct experiments on three multi-turn agent benchmarks:
ALFWorld~\cite{alfworld}, WebShop~\cite{webshop}, and
ScienceWorld~\cite{scienceworld}. We follow TCOD~\cite{tcod} in using the same data splits, environment configurations, and prompt templates. The maximum numbers of environment interaction steps are set to 30, 15, and 30 for ALFWorld, WebShop, and ScienceWorld, respectively. Our main experiments use Qwen3-32B as the teacher and we further evaluate Qwen3-4B as the student and a Qwen3-8B teacher trained with reinforcement learning~\cite{qwen3,grpo,gigpo}. Further details of the teacher training are provided in Appendix~\ref{sec:appendix_rl_teacher}. We report success rate for ALFWorld, and both task score and success rate for WebShop and ScienceWorld.

\noindent\textbf{Training Setup.}
All methods use the same training budget of 200 optimization steps. FTB is built upon the B2F curriculum introduced by TCOD~\cite{tcod}, where the teacher prefix is shortened by one environment step every five training steps, allowing the student to progressively take control from earlier states. Further experimental details are provided in Appendix~\ref{sec:appendix_implementation}.

\noindent\textbf{Baselines.}
We compare with several agentic on-policy distillation methods. Vanilla OPD directly performs teacher--student distillation on states visited by the student~\cite{agarwal2024onpolicy}. TCOD-F2B starts from the early part of the trajectory and progressively increases the student rollout depth, while TCOD-B2F initializes the student from intermediate states using prefixes of successful reference trajectories and gradually shortens these prefixes during training~\cite{tcod}. Guided-OPD mixes teacher and student turns through a curriculum schedule, progressively increasing the student's control over the full trajectory~\cite{guided-opd}. All baselines are implemented following the core settings of their original papers.

\subsection{Main Results}

\noindent\textbf{Results with the Qwen3-32B Teacher.}
As shown in Table~\ref{tab:main_results}, with the Qwen3-1.7B student, FTB improves the average success rate over OPD and TCOD-B2F by 16.6 and 7.6 percentage points, respectively. When scaling the student to Qwen3-4B, FTB still yields improvements of 7.5 and 5.2 percentage points, demonstrating consistent gains across student scales. Notably, on WebShop, the 1.7B student trained with FTB outperforms both the 4B student and the zero-shot 32B teacher. Trajectory analysis shows that the 1.7B student follows simpler and more direct paths, averaging 7.67 steps with a 32\% maximum step rate, while the 4B student averages 11.37 steps with a 51\% maximum step rate and more often becomes trapped in redundant interactions. FTB trains the student on validated local guidance and supports the development of a task specific policy with more efficient execution, helping the student surpass the zero-shot teacher.

\begin{table}[!t]
\centering
\footnotesize
\setlength{\tabcolsep}{1.8pt}

\begin{tabular*}{\columnwidth}{@{\extracolsep{\fill}}lccc}
\toprule
\multirow{2}{*}{\textbf{Method}}
& \textbf{ALFWorld}
& \multicolumn{2}{c}{\textbf{WebShop}} \\
\cmidrule(lr){2-2}
\cmidrule(lr){3-4}
& \textbf{SR $\uparrow$}
& \textbf{Score $\uparrow$}
& \textbf{SR $\uparrow$} \\
\midrule

\rowcolor{groupGray}
\multicolumn{4}{c}{
\textnormal{Qwen3-8B-RL teacher $\rightarrow$ Qwen3-4B student}
}
\\

Student (zero-shot)
& 20.9
& 32.9
& 29.0
\\
\addlinespace[1pt]

Teacher (zero-shot)
& 79.9
& 62.6
& 61.0
\\

\addlinespace[1pt]

OPD
& 67.4 $\pm$ 3.7
& 63.7 $\pm$ 2.5
& 64.0 $\pm$ 1.6
\\

TCOD-F2B
& 66.5 $\pm$ 1.3
& 63.3 $\pm$ 2.6
& \secondcell{64.7 $\pm$ 2.1}
\\

TCOD-B2F
& 68.6 $\pm$ 2.3
& 61.7 $\pm$ 1.7
& 63.3 $\pm$ 1.2
\\

Guided-OPD
& \secondcell{72.6 $\pm$ 3.4}
& \secondcell{63.8 $\pm$ 3.9}
& 62.0 $\pm$ 5.0
\\

FTB w/o Bridge Exec.
& 70.8 $\pm$ 1.6
& 63.0 $\pm$ 0.8
& 63.0 $\pm$ 1.4
\\

\textbf{FTB (Ours)}
& \topcell{74.1 $\pm$ 3.1}
& \topcell{65.7 $\pm$ 1.2}
& \topcell{66.3 $\pm$ 0.5}
\\

\bottomrule
\end{tabular*}

\caption{Results with an RL-trained teacher, reported as mean $\pm$ standard deviation over three seeds; \textbf{blue} and \textbf{red} backgrounds denote the best and second-best trained methods for each reported metric, respectively.}
\label{tab:rl_teacher_results}
\end{table}

\noindent\textbf{Results with a Teacher Trained with RL.}
As shown in Table~\ref{tab:rl_teacher_results}, when using the Qwen3-8B teacher trained with RL, FTB achieves the best performance on all three reported metrics. Its average success rate across ALFWorld and WebShop exceeds OPD and TCOD-B2F by 4.5 and 4.3 percentage points, respectively, confirming its effectiveness under different teacher configurations considered here.

\begin{figure*}[!t]
    \centering
    \includegraphics[width=0.98\textwidth]{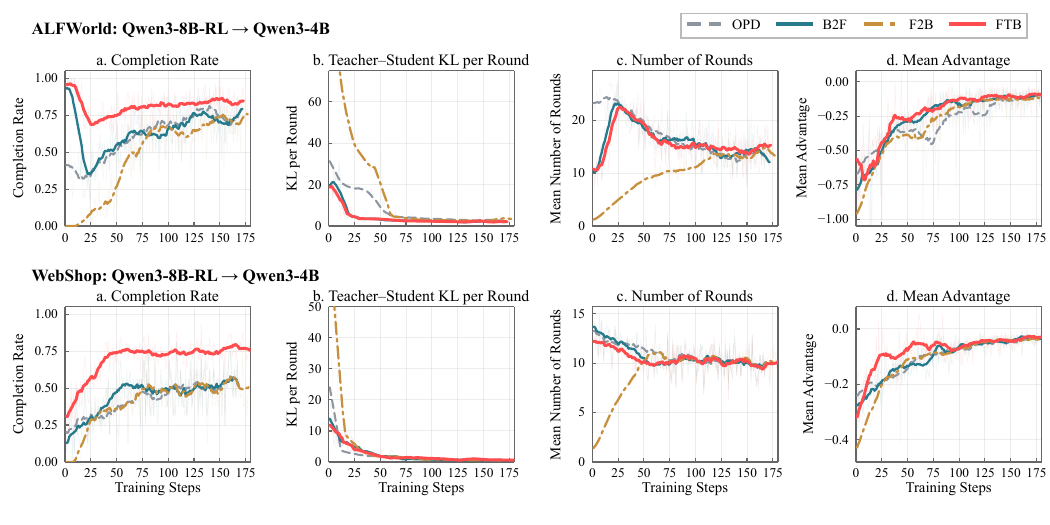}
    \caption{Training dynamics on ALFWorld and WebShop under the Qwen3-8B-RL teacher $\rightarrow$ Qwen3-4B student setting. The top and bottom rows correspond to ALFWorld and WebShop, respectively. From left to right, the four columns report the trajectory completion rate, teacher--student KL divergence per interaction round, mean number of interaction rounds, and mean sampled distillation advantage over training.}
\label{fig:training_dynamics}
\end{figure*}

\noindent\textbf{Training Dynamics.} Figure~\ref{fig:training_dynamics} compares the training dynamics under the Qwen3-8B-RL teacher $\rightarrow$ Qwen3-4B student setting. On both benchmarks, FTB improves the completion rate earlier and rapidly reduces the teacher--student KL difference. As training progresses, the numbers of interaction rounds gradually converge to similar levels across methods, while FTB achieves a faster increase in mean distillation advantage, indicating more effective optimization of the student policy during training.

\subsection{Ablation Study}

\begin{table}[!t]
\centering
\footnotesize
\setlength{\tabcolsep}{1.8pt}

\begin{tabular*}{\columnwidth}{
    @{\extracolsep{\fill}}
    l
    c
    cc
}
\toprule

\multirow{2}{*}{\textbf{Method}}
&
\textbf{ALFWorld}
&
\multicolumn{2}{c}{\textbf{WebShop}}
\\

\cmidrule(lr){2-2}
\cmidrule(lr){3-4}

&
\textbf{SR $\uparrow$}
&
\textbf{Score $\uparrow$}
&
\textbf{SR $\uparrow$}
\\

\midrule

\rowcolor{groupGray}
\multicolumn{4}{c}{
\textnormal{Qwen3-8B-RL teacher $\rightarrow$ Qwen3-4B student}
}
\\

Student (zero-shot)
& 20.9
& 32.9
& 29.0
\\

Teacher (zero-shot)
& 79.9
& 62.6
& 61.0
\\

\addlinespace[1pt]

TCOD-B2F
& 68.6 $\pm$ 2.3
& 61.7 $\pm$ 1.7
& 63.3 $\pm$ 1.2
\\

\addlinespace[2pt]

FTB w/ Random Turn
& 62.3 $\pm$ 3.0
& 58.3 $\pm$ 1.3
& 60.3 $\pm$ 1.2
\\

FTB w/o Bridge Exec.
& 70.8 $\pm$ 1.6
& 63.0 $\pm$ 0.8
& 63.0 $\pm$ 1.4
\\

FTB w/o Future Validation
& \secondcell{71.9 $\pm$ 5.8}
& \secondcell{65.0 $\pm$ 1.1}
& \secondcell{64.3 $\pm$ 3.2}
\\

\textbf{FTB (Full)}
& \topcell{74.1 $\pm$ 3.1}
& \topcell{65.7 $\pm$ 1.2}
& \topcell{66.3 $\pm$ 0.5}
\\

\bottomrule
\end{tabular*}

\caption{Ablation results with an RL-trained teacher, reported as mean $\pm$ standard deviation over three seeds; \textbf{blue} and \textbf{red} backgrounds denote the best and second-best trained methods for each reported metric, respectively.}
\label{tab:ablation_results}

\end{table}

Following the Qwen3-8B-RL teacher $\rightarrow$ Qwen3-4B student setting in Table~\ref{tab:rl_teacher_results}, we conduct the ablation study in Table~\ref{tab:ablation_results} to examine candidate position selection, teacher bridge execution, and future trajectory validation. \textit{FTB w/ Random Turn} replaces the turn with the largest teacher--student disagreement with a randomly selected turn, evaluating the importance of locating critical intervention positions. \textit{FTB w/o Bridge Exec.} does not execute the teacher bridge in the environment. It determines whether to include the corresponding teacher action in training by comparing the teacher preferred token ratios of the trajectory segments before and after the candidate position in the original student trajectory. \textit{FTB w/o Future Validation} executes the teacher bridge and lets the student continue from the resulting state, but no longer filters the teacher action according to whether the subsequent trajectory achieves a higher teacher preferred token ratio than the original trajectory. Random turn selection causes the largest performance degradation, while removing bridge execution or future validation also consistently reduces performance. The full FTB achieves the best results across all metrics.

\section{Analysis}

\begin{figure}[t]
    \centering
    \includegraphics[width=\columnwidth]{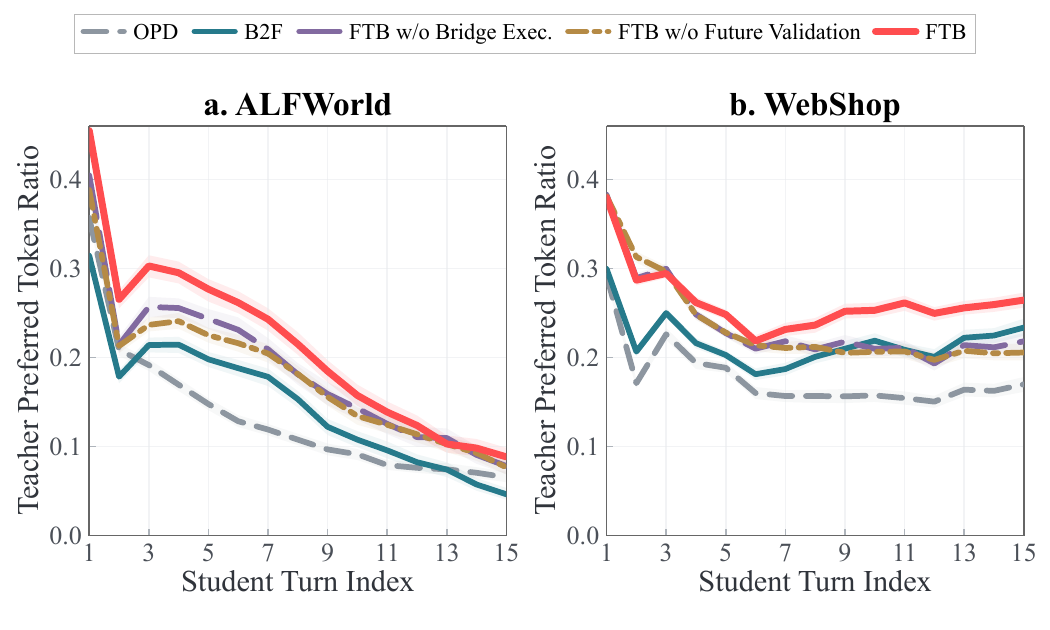}
    \caption{Teacher preferred token ratio over student turns on ALFWorld and WebShop.}
\label{fig:teacher_alignment}
\end{figure}

\noindent\textbf{Teacher Preferred Token Ratio over Student Turns.}
We measure the teacher preferred token ratio at each turn after the student takes control. As shown in Figure~\ref{fig:teacher_alignment}, FTB achieves a higher teacher preferred token ratio over most turns on both ALFWorld and WebShop, whereas removing bridge execution or future validation generally degrades it. Moreover, although bridge tokens constitute only 8.8\% of the tokens used for optimization, they contribute 75.2\% of the total positive distillation advantage. These results suggest that future trajectory validation identifies a small set of informative teacher actions that improve the alignment of subsequent student trajectories with the teacher.

\noindent\textbf{Bridge Characteristics.}
We analyze retained teacher bridges in terms of their trigger positions, disagreement dynamics, and modifications to student actions.

\begin{figure}[t]
    \centering
    \includegraphics[width=\columnwidth]{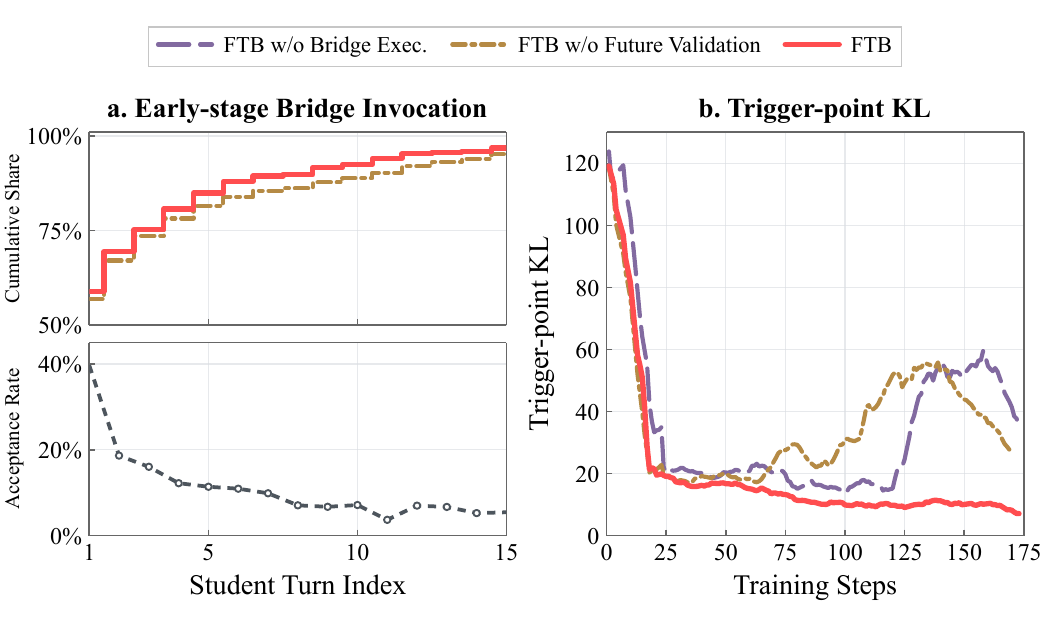}
    \caption{Bridge invocation, acceptance, and trigger point disagreement during training.}
\label{fig:bridge_analysis}
\end{figure}

\textbf{Trigger Positions and Acceptance Rates.} Figure~\ref{fig:bridge_analysis} (a) shows the cumulative distribution of retained bridges over student turns and the fraction of candidates with high disagreement accepted by the future validation gate. More than half of the bridges occur at the first student turn, and approximately 85\% occur within the first five turns. The acceptance rate also decreases over turns, from 39.7\% at the first turn to 18.6\% at the second and approximately 4--7\% at later turns. Early positions are therefore selected more often and are more likely to yield improved student continuations. This is consistent with our earlier observation that early deviations have stronger effects on subsequent interactions.

\textbf{Disagreement at Trigger Positions.} Figure~\ref{fig:bridge_analysis}(b) shows the KL divergence between the teacher and student at the selected trigger positions over training. Under full FTB, the KL divergence decreases rapidly and remains low. Removing bridge execution or future validation causes it to rebound during later training. Local disagreement can identify positions that may benefit from correction, but cannot determine by itself whether a teacher action is useful. Bridge execution and future validation further remove candidates that fail to improve the subsequent trajectory.

\textbf{Modifications to Student Actions.} Retained bridges are rarely identical to the original student actions. In ALFWorld, 91.5\% of bridges modify the action. Among them, 58.9\% preserve the action type but change its target or argument, while 32.7\% change the action type. In WebShop, 68.9\% of bridges modify the action. Among them, 64.2\% adjust the arguments while preserving the search or click type, and only 4.8\% change the action type. These patterns show that retained bridges can either alter the action structure or preserve the high level operation while making fine grained corrections to the student policy.

\noindent\textbf{Training Efficiency.}
Table~\ref{tab:training_efficiency} reports the average runtime per step on ALFWorld. Despite performing an additional teacher bridge and a short student validation continuation, FTB is consistently faster than OPD across student scales, achieving $0.86\times$ and $0.64\times$ of OPD's runtime for the 1.7B and 4B students, respectively. The reason is that OPD requires a full student rollout at every training step and thus scales strongly with student size, whereas FTB inherits the shortened student rollouts of the B2F curriculum and only adds bounded validation overhead on top. Compared with B2F, FTB introduces roughly $120$s and $77$s of additional runtime on the 1.7B and 4B students, respectively, indicating that the validation overhead remains moderate and does not grow proportionally with student size. Overall, future validation adds no substantial training-time penalty, and its efficiency advantage over OPD grows with student size.

\begin{table}[t]
\centering
\small
\setlength{\tabcolsep}{5.5pt}
\renewcommand{\arraystretch}{1.06}
\resizebox{\columnwidth}{!}{%
\begin{tabular}{lcc}
\toprule
\multirow{2}{*}{\textbf{Method}}
& \multicolumn{2}{c}{\textbf{Qwen3-32B teacher}} \\
\cmidrule(lr){2-3}
& \textbf{Qwen3-1.7B student}
& \textbf{Qwen3-4B student} \\
\midrule
OPD
& 481.5 ($1.00\times$)
& 908.2 ($1.00\times$) \\
TCOD-B2F
& 296.3 ($0.62\times$)
& 500.8 ($0.55\times$) \\
TCOD-F2B
& 193.1 ($0.40\times$)
& 212.2 ($0.23\times$) \\
\textbf{FTB (Ours)}
& \textbf{416.7 ($0.86\times$)}
& \textbf{577.9 ($0.64\times$)} \\
\bottomrule
\end{tabular}%
}
\caption{Training time per step on ALFWorld, averaged over 200 steps under the same 8$\times$A100 setup. Relative costs with respect to OPD are shown in parentheses.}
\label{tab:training_efficiency}
\end{table}

\section{Conclusion}

We propose FutureBridge-OPD (FTB) to mitigate error accumulation in multi-turn agentic on-policy distillation by identifying high-disagreement states, executing local teacher bridges, and validating them through induced student continuations. Unlike methods that assess guidance using local teacher--student disagreement, FTB explicitly evaluates whether an intervention redirects future student behavior toward regions with stronger teacher preference. Experiments on ALFWorld, WebShop, and ScienceWorld demonstrate consistent gains across student scales and teacher configurations. Ablations and trajectory analyses further confirm the complementary roles of candidate localization, bridge execution, and future validation. These results establish future trajectory validation as an effective way to select useful guidance in agentic distillation.

\bibliographystyle{plainnat}
\bibliography{references}

@inproceedings{agarwal2024onpolicy,
  author       = {Rishabh Agarwal and
                  Nino Vieillard and
                  Yongchao Zhou and
                  Piotr Stanczyk and
                  Sabela Ramos Garea and
                  Matthieu Geist and
                  Olivier Bachem},
  title        = {On-Policy Distillation of Language Models: Learning from Self-Generated
                  Mistakes},
  booktitle    = {The Twelfth International Conference on Learning Representations,
                  {ICLR} 2024, Vienna, Austria, May 7-11, 2024},
  publisher    = {OpenReview.net},
  year         = {2024},
  url          = {https://openreview.net/forum?id=3zKtaqxLhW},
  bibsource    = {dblp computer science bibliography, https://dblp.org}
}

@misc{tcod,
      title={TCOD: Exploring Temporal Curriculum in On-Policy Distillation for Multi-turn Autonomous Agents}, 
      author={Jiaqi Wang and Wenhao Zhang and Weijie Shi and Yaliang Li and James Cheng},
      year={2026},
      eprint={2604.24005},
      archivePrefix={arXiv},
      primaryClass={cs.LG},
      url={https://arxiv.org/abs/2604.24005}, 
}

@misc{guided-opd,
      title={On-Policy Distillation with Curriculum Turn-level Guidance for Multi-turn Agents}, 
      author={Gengsheng Li and Mao Zheng and Mingyang Song and Ruiqi Liu and Tianyu Yang and Jie Sun and Qiyong Zhong and Haiyun Guo and Junfeng Fang and Dan Zhang and Jinqiao Wang},
      year={2026},
      eprint={2606.15912},
      archivePrefix={arXiv},
      primaryClass={cs.LG},
      url={https://arxiv.org/abs/2606.15912}, 
}

@misc{turnopd,
      title={TurnOPD: Making On-Policy Distillation Turn-Aware for Efficient Long-Horizon Agent Training}, 
      author={Yuhang Zhou and Kai Zheng and Haoling Li and Dengyun Peng and Can Xu and Jingjing Chen},
      year={2026},
      eprint={2607.05804},
      archivePrefix={arXiv},
      primaryClass={cs.AI},
      url={https://arxiv.org/abs/2607.05804}, 
}

@misc{TIP,
      title={TIP: Token Importance in On-Policy Distillation}, 
      author={Yuanda Xu and Hejian Sang and Zhengze Zhou and Ran He and Zhipeng Wang and Alborz Geramifard},
      year={2026},
      eprint={2604.14084},
      archivePrefix={arXiv},
      primaryClass={cs.LG},
      url={https://arxiv.org/abs/2604.14084}, 
}

@misc{SCOPE,
      title={SCOPE: Signal-Calibrated On-Policy Distillation Enhancement with Dual-Path Adaptive Weighting}, 
      author={Binbin Zheng and Xing Ma and Yiheng Liang and Jingqing Ruan and Xiaoliang Fu and Kepeng Lin and Benchang Zhu and Ke Zeng and Xunliang Cai},
      year={2026},
      eprint={2604.10688},
      archivePrefix={arXiv},
      primaryClass={cs.LG},
      url={https://arxiv.org/abs/2604.10688}, 
}

@misc{SAGE-OPD,
      title={SAGE-OPD: Selective Agent-Guided Intervention for Multi-Turn On-Policy Distillation}, 
      author={Yuhang Zhou and Lizhu Zhang and Yifan Wu and Mingyi Wang and Bo Peng and Jiayi Liu and Xiangjun Fan and Zhuokai Zhao},
      year={2026},
      eprint={2606.19659},
      archivePrefix={arXiv},
      primaryClass={cs.CL},
      url={https://arxiv.org/abs/2606.19659}, 
}

@misc{ATOD,
      title={ATOD: Annealed Turn-aware On-policy Distillation for Multi-turn Autonomous Agents}, 
      author={Qitai Tan and Zefang Zong and Mo Li and Yipeng Shi and Yang Li and Peng Chen},
      year={2026},
      eprint={2606.27814},
      archivePrefix={arXiv},
      primaryClass={cs.AI},
      url={https://arxiv.org/abs/2606.27814}, 
}

@misc{FIPO,
      title={FIPO: Eliciting Deep Reasoning with Future-KL Influenced Policy Optimization}, 
      author={Chiyu Ma and Shuo Yang and Kexin Huang and Jinda Lu and Haoming Meng and Shangshang Wang and Bolin Ding and Soroush Vosoughi and Guoyin Wang and Jingren Zhou},
      year={2026},
      eprint={2603.19835},
      archivePrefix={arXiv},
      primaryClass={cs.LG},
      url={https://arxiv.org/abs/2603.19835}, 
}

@misc{APPO,
      title={APPO: Agentic Procedural Policy Optimization}, 
      author={Xucong Wang and Ziyu Ma and Yong Wang and Yuxiang Ji and Shidong Yang and Guanhua Chen and Pengkun Wang and Xiangxiang Chu},
      year={2026},
      eprint={2606.12384},
      archivePrefix={arXiv},
      primaryClass={cs.LG},
      url={https://arxiv.org/abs/2606.12384}, 
}

@misc{reopd,
      title={Multi-Turn On-Policy Distillation with Prefix Replay}, 
      author={Baohao Liao and Hanze Dong and Christof Monz and Xinxing Xu and Li Dong and Furu Wei},
      year={2026},
      eprint={2607.04763},
      archivePrefix={arXiv},
      primaryClass={cs.LG},
      url={https://arxiv.org/abs/2607.04763}, 
}

@inproceedings{prefixopd,
  author       = {Dongxu Zhang and
                  Zhichao Yang and
                  Sepehr Janghorbani and
                  Jun Han and
                  Andrew Ressler II and
                  Qian Qian and
                  Gregory D. Lyng and
                  Sanjit Singh Batra and
                  Robert E. Tillman},
  editor       = {Maria Liakata and
                  Viviane P. Moreira and
                  Jiajun Zhang and
                  David Jurgens},
  title        = {Fast and Effective On-Policy Distillation from Reasoning Prefixes},
  booktitle    = {Findings of the Association for Computational Linguistics, {ACL} 2026,
                  San Diego, California, United States, July 2-7, 2026},
  pages        = {25553--25569},
  publisher    = {Association for Computational Linguistics},
  year         = {2026},
  url          = {https://aclanthology.org/2026.findings-acl.1276/},
  bibsource    = {dblp computer science bibliography, https://dblp.org}
}

@inproceedings{
eopd,
title={Entropy-Aware On-Policy Distillation of Language Models},
author={Woogyeol Jin and Taywon Min and Yongjin Yang and Dennis Wei and Yi Zhou and Swanand Ravindra Kadhe and Nathalie Baracaldo and Kimin Lee},
booktitle={Forty-third International Conference on Machine Learning},
year={2026},
url={https://openreview.net/forum?id=J5i09faOOf}
}

@misc{egrsd,
      title={Respecting Self-Uncertainty in On-Policy Self-Distillation for Efficient LLM Reasoning}, 
      author={Junlong Ke and Zichen Wen and Weijia Li and Conghui He and Linfeng Zhang},
      year={2026},
      eprint={2605.13255},
      archivePrefix={arXiv},
      primaryClass={cs.AI},
      url={https://arxiv.org/abs/2605.13255}, 
}

@misc{sod,
      title={SOD: Step-wise On-policy Distillation for Small Language Model Agents}, 
      author={Qiyong Zhong and Mao Zheng and Mingyang Song and Xin Lin and Jie Sun and Houcheng Jiang and Xiang Wang and Junfeng Fang},
      year={2026},
      eprint={2605.07725},
      archivePrefix={arXiv},
      primaryClass={cs.CL},
      url={https://arxiv.org/abs/2605.07725}, 
}

@inproceedings{oar,
  author       = {Ziheng Li and
                  Liu Kang and
                  Feng Xiao and
                  Luxi Xing and
                  Qingyi Si and
                  Zhuoran Li and
                  Weikang Gong and
                  Deqing Yang and
                  Yanghua Xiao and
                  Hongcheng Guo},
  editor       = {Maria Liakata and
                  Viviane P. Moreira and
                  Jiajun Zhang and
                  David Jurgens},
  title        = {Outcome-Grounded Advantage Reshaping for Fine-Grained Credit Assignment
                  in Mathematical Reasoning},
  booktitle    = {Proceedings of the 64th Annual Meeting of the Association for Computational
                  Linguistics (Volume 1: Long Papers), {ACL} 2026, San Diego, California,
                  United States, July 2-7, 2026},
  pages        = {24681--24693},
  publisher    = {Association for Computational Linguistics},
  year         = {2026},
  url          = {https://aclanthology.org/2026.acl-long.1132/},
  bibsource    = {dblp computer science bibliography, https://dblp.org}
}

@misc{rrpo,
      title={Credit Assignment with Resets in Language Model Reasoning}, 
      author={Ankur Samanta and Akshayaa Magesh and Ayush Jain and Youliang Yu and Daniel Jiang and Kavosh Asadi and Kaveh Hassani and Paul Sajda and Jalaj Bhandari and Yonathan Efroni},
      year={2026},
      eprint={2605.25507},
      archivePrefix={arXiv},
      primaryClass={cs.AI},
      url={https://arxiv.org/abs/2605.25507}, 
}

@inproceedings{alfworld,
  author       = {Mohit Shridhar and
                  Xingdi Yuan and
                  Marc{-}Alexandre C{\^{o}}t{\'{e}} and
                  Yonatan Bisk and
                  Adam Trischler and
                  Matthew J. Hausknecht},
  title        = {ALFWorld: Aligning Text and Embodied Environments for Interactive
                  Learning},
  booktitle    = {9th International Conference on Learning Representations, {ICLR} 2021,
                  Virtual Event, Austria, May 3-7, 2021},
  publisher    = {OpenReview.net},
  year         = {2021},
  url          = {https://openreview.net/forum?id=0IOX0YcCdTn},
  bibsource    = {dblp computer science bibliography, https://dblp.org}
}

@inproceedings{webshop,
  author       = {Shunyu Yao and
                  Howard Chen and
                  John Yang and
                  Karthik Narasimhan},
  editor       = {Sanmi Koyejo and
                  S. Mohamed and
                  A. Agarwal and
                  Danielle Belgrave and
                  K. Cho and
                  A. Oh},
  title        = {WebShop: Towards Scalable Real-World Web Interaction with Grounded
                  Language Agents},
  booktitle    = {Advances in Neural Information Processing Systems 35: Annual Conference
                  on Neural Information Processing Systems 2022, NeurIPS 2022, New Orleans,
                  LA, USA, November 28 - December 9, 2022},
  year         = {2022},
  url          = {http://papers.nips.cc/paper\_files/paper/2022/hash/82ad13ec01f9fe44c01cb91814fd7b8c-Abstract-Conference.html},
  bibsource    = {dblp computer science bibliography, https://dblp.org}
}

@misc{qwen3,
      title={Qwen3 Technical Report}, 
      author={An Yang and Anfeng Li and Baosong Yang and Beichen Zhang and Binyuan Hui and Bo Zheng and Bowen Yu and Chang Gao and Chengen Huang and Chenxu Lv and Chujie Zheng and Dayiheng Liu and Fan Zhou and Fei Huang and Feng Hu and Hao Ge and Haoran Wei and Huan Lin and Jialong Tang and Jian Yang and Jianhong Tu and Jianwei Zhang and Jianxin Yang and Jiaxi Yang and Jing Zhou and Jingren Zhou and Junyang Lin and Kai Dang and Keqin Bao and Kexin Yang and Le Yu and Lianghao Deng and Mei Li and Mingfeng Xue and Mingze Li and Pei Zhang and Peng Wang and Qin Zhu and Rui Men and Ruize Gao and Shixuan Liu and Shuang Luo and Tianhao Li and Tianyi Tang and Wenbiao Yin and Xingzhang Ren and Xinyu Wang and Xinyu Zhang and Xuancheng Ren and Yang Fan and Yang Su and Yichang Zhang and Yinger Zhang and Yu Wan and Yuqiong Liu and Zekun Wang and Zeyu Cui and Zhenru Zhang and Zhipeng Zhou and Zihan Qiu},
      year={2025},
      eprint={2505.09388},
      archivePrefix={arXiv},
      primaryClass={cs.CL},
      url={https://arxiv.org/abs/2505.09388}, 
}

@inproceedings{gigpo,
 author = {Feng, Lang and Xue, Zhenghai and Liu, Tingcong and An, Bo},
 booktitle = {Advances in Neural Information Processing Systems},
 editor = {D. Belgrave and C. Zhang and H. Lin and R. Pascanu and P. Koniusz and M. Ghassemi and N. Chen},
 pages = {46375--46408},
 publisher = {Curran Associates, Inc.},
 title = {Group-in-Group Policy Optimization for LLM Agent Training},
 url = {https://proceedings.neurips.cc/paper_files/paper/2025/file/420c9f777c0b4f78d515e53cf74d58b2-Paper-Conference.pdf},
 volume = {38},
 year = {2025}
}

@misc{grpo,
      title={DeepSeekMath: Pushing the Limits of Mathematical Reasoning in Open Language Models}, 
      author={Zhihong Shao and Peiyi Wang and Qihao Zhu and Runxin Xu and Junxiao Song and Xiao Bi and Haowei Zhang and Mingchuan Zhang and Y. K. Li and Y. Wu and Daya Guo},
      year={2024},
      eprint={2402.03300},
      archivePrefix={arXiv},
      primaryClass={cs.CL},
      url={https://arxiv.org/abs/2402.03300}, 
}

@inproceedings{scienceworld,
  author       = {Ruoyao Wang and
                  Peter A. Jansen and
                  Marc{-}Alexandre C{\^{o}}t{\'{e}} and
                  Prithviraj Ammanabrolu},
  editor       = {Yoav Goldberg and
                  Zornitsa Kozareva and
                  Yue Zhang},
  title        = {ScienceWorld: Is your Agent Smarter than a 5th Grader?},
  booktitle    = {Proceedings of the 2022 Conference on Empirical Methods in Natural
                  Language Processing, {EMNLP} 2022, Abu Dhabi, United Arab Emirates,
                  December 7-11, 2022},
  pages        = {11279--11298},
  publisher    = {Association for Computational Linguistics},
  year         = {2022},
  url          = {https://doi.org/10.18653/v1/2022.emnlp-main.775},
  doi          = {10.18653/V1/2022.EMNLP-MAIN.775},
  bibsource    = {dblp computer science bibliography, https://dblp.org}
}

@misc{shortopd,
      title={ShortOPD: Recovering Pruned LLMs with Short-to-Long On-Policy Distillation}, 
      author={Qingyu Zhang and Qianhao Yuan and Hongyu Lin and Yaojie Lu and Xianpei Han and Le Sun and Xiang Li and Ming Xu and Jiarui Li and Xiuyin Zhao},
      year={2026},
      eprint={2607.13124},
      archivePrefix={arXiv},
      primaryClass={cs.LG},
      url={https://arxiv.org/abs/2607.13124}, 
}

@misc{topd,
      title={Bridging Reasoning Trajectories in On-Policy Distillation via Near-Future Guidance}, 
      author={Yuxuan Jiang and Francis Ferraro},
      year={2026},
      eprint={2606.00305},
      archivePrefix={arXiv},
      primaryClass={cs.CL},
      url={https://arxiv.org/abs/2606.00305}, 
}

@misc{rethinkopd,
      title={Rethinking On-Policy Distillation of Large Language Models: Phenomenology, Mechanism, and Recipe}, 
      author={Yaxuan Li and Yuxin Zuo and Bingxiang He and Jinqian Zhang and Chaojun Xiao and Cheng Qian and Tianyu Yu and Huan-ang Gao and Wenkai Yang and Zhiyuan Liu and Ning Ding},
      year={2026},
      eprint={2604.13016},
      archivePrefix={arXiv},
      primaryClass={cs.LG},
      url={https://arxiv.org/abs/2604.13016}, 
}

@misc{yourteacher,
      title={Your Teacher Can't Help You Here: Combating Supervision Fidelity Decay in On-Policy Distillation}, 
      author={Yanjiang Liu and Jie Lou and Xinyan Guan and Yuqiu Ji and Hongyu Lin and Ben He and Xianpei Han and Le Sun and Xing Yu and Yaojie Lu},
      year={2026},
      eprint={2605.30833},
      archivePrefix={arXiv},
      primaryClass={cs.CL},
      url={https://arxiv.org/abs/2605.30833}, 
}

@misc{pruneopd,
      title={Prune-OPD: Efficient and Reliable On-Policy Distillation for Long-Horizon Reasoning}, 
      author={Zhicheng Yang and Zhijiang Guo and Yifan Song and Minrui Xu and Yongxin Wang and Yiwei Wang and Xiaodan Liang and Jing Tang},
      year={2026},
      eprint={2605.07804},
      archivePrefix={arXiv},
      primaryClass={cs.LG},
      url={https://arxiv.org/abs/2605.07804}, 
}

@misc{distillingtheknowledge,
      title={Distilling the Knowledge in a Neural Network}, 
      author={Geoffrey Hinton and Oriol Vinyals and Jeff Dean},
      year={2015},
      eprint={1503.02531},
      archivePrefix={arXiv},
      primaryClass={stat.ML},
      url={https://arxiv.org/abs/1503.02531}, 
}

@inproceedings{minillm,
  author       = {Yuxian Gu and
                  Li Dong and
                  Furu Wei and
                  Minlie Huang},
  title        = {MiniLLM: Knowledge Distillation of Large Language Models},
  booktitle    = {The Twelfth International Conference on Learning Representations,
                  {ICLR} 2024, Vienna, Austria, May 7-11, 2024},
  publisher    = {OpenReview.net},
  year         = {2024},
  url          = {https://openreview.net/forum?id=5h0qf7IBZZ},
  bibsource    = {dblp computer science bibliography, https://dblp.org}
}

@misc{RevisitingOn-PolicyDistillation,
      title={Revisiting On-Policy Distillation: Empirical Failure Modes and Simple Fixes}, 
      author={Yuqian Fu and Haohuan Huang and Kaiwen Jiang and Jiacai Liu and Zhuo Jiang and Yuanheng Zhu and Dongbin Zhao},
      year={2026},
      eprint={2603.25562},
      archivePrefix={arXiv},
      primaryClass={cs.LG},
      url={https://arxiv.org/abs/2603.25562}, 
}

@inproceedings{AReductionofImitation,
  author       = {St{\'{e}}phane Ross and
                  Geoffrey J. Gordon and
                  Drew Bagnell},
  editor       = {Geoffrey J. Gordon and
                  David B. Dunson and
                  Miroslav Dud{\'{\i}}k},
  title        = {A Reduction of Imitation Learning and Structured Prediction to No-Regret
                  Online Learning},
  booktitle    = {Proceedings of the Fourteenth International Conference on Artificial
                  Intelligence and Statistics, {AISTATS} 2011, Fort Lauderdale, USA,
                  April 11-13, 2011},
  series       = {{JMLR} Proceedings},
  volume       = {15},
  pages        = {627--635},
  publisher    = {JMLR.org},
  year         = {2011},
  url          = {http://proceedings.mlr.press/v15/ross11a/ross11a.pdf},
  bibsource    = {dblp computer science bibliography, https://dblp.org}
}

@inproceedings{ScheduledSampling,
  author       = {Samy Bengio and
                  Oriol Vinyals and
                  Navdeep Jaitly and
                  Noam Shazeer},
  editor       = {Corinna Cortes and
                  Neil D. Lawrence and
                  Daniel D. Lee and
                  Masashi Sugiyama and
                  Roman Garnett},
  title        = {Scheduled Sampling for Sequence Prediction with Recurrent Neural Networks},
  booktitle    = {Advances in Neural Information Processing Systems 28: Annual Conference
                  on Neural Information Processing Systems 2015, December 7-12, 2015,
                  Montreal, Quebec, Canada},
  pages        = {1171--1179},
  year         = {2015},
  url          = {https://proceedings.neurips.cc/paper/2015/hash/e995f98d56967d946471af29d7bf99f1-Abstract.html},
  bibsource    = {dblp computer science bibliography, https://dblp.org}
}

@inproceedings{Sequence-Level-knowledge,
  author       = {Yoon Kim and
                  Alexander M. Rush},
  editor       = {Jian Su and
                  Xavier Carreras and
                  Kevin Duh},
  title        = {Sequence-Level Knowledge Distillation},
  booktitle    = {Proceedings of the 2016 Conference on Empirical Methods in Natural
                  Language Processing, {EMNLP} 2016, Austin, Texas, USA, November 1-4,
                  2016},
  pages        = {1317--1327},
  publisher    = {The Association for Computational Linguistics},
  year         = {2016},
  url          = {https://doi.org/10.18653/v1/d16-1139},
  doi          = {10.18653/V1/D16-1139},
  bibsource    = {dblp computer science bibliography, https://dblp.org}
}

@misc{tropd,
      title={Trust Region On-Policy Distillation}, 
      author={Xingrun Xing and Haoqing Wang and Boyan Gao and Ziheng Li and Yehui Tang},
      year={2026},
      eprint={2606.01249},
      archivePrefix={arXiv},
      primaryClass={cs.LG},
      url={https://arxiv.org/abs/2606.01249}, 
}

@misc{relayopd,
      title={Pass the Baton: Trajectory-Relayed On-Policy Distillation}, 
      author={Haolei Xu and Xiaowen Xu and Haiwen Hong and Zixuan Ni and Hongxing Li and Yiwen Qiu and Weiming Lu and Yongliang Shen},
      year={2026},
      eprint={2607.26057},
      archivePrefix={arXiv},
      primaryClass={cs.CL},
      url={https://arxiv.org/abs/2607.26057}, 
}

\clearpage
\appendix

\section{Overview}

This supplement reports the intervention analysis motivating FTB,
experimental settings and additional implementation details, formal
results for sampled-token reverse-KL estimation and future-trajectory
validation, sensitivity to the validation horizon, distinctions from
closely related methods, and limitations.

\section{Motivating Intervention Analysis}
\label{sec:appendix_motivation}

We analyze 1,000 trajectories that are unsuccessful according to the
terminal task-success criterion of each benchmark, comprising 500
ALFWorld, 250 WebShop, and 250 ScienceWorld trajectories.

For each trajectory, we identify the eligible Student-controlled turn
with the largest Teacher--Student disagreement, replace the original
Student response at that turn with a Teacher response, and then let the
same frozen Student policy complete the remaining interaction.

Because the available outcome signals differ across benchmarks, we use
a benchmark-specific comparison rule. For ALFWorld, only binary
terminal task success is available for this analysis. An intervention
is therefore classified as improved if it converts the originally
failed episode into a successful one and unchanged otherwise; a
degraded outcome is not applicable under this binary comparison. For
WebShop and ScienceWorld, which additionally provide graded process
rewards, we compare the benchmark-specific trajectory-level process
score after intervention with that of the original trajectory and
classify the intervention as improved, unchanged, or degraded.

For the random baseline, the intervention position is sampled uniformly
from the same set of eligible Student-controlled turns and evaluated
using the same benchmark-specific comparison rule.

As shown in Table~\ref{tab:motivating_intervention}, across all 1,000
trajectories, interventions at high-disagreement positions improve 309
trajectories (30.9\%), compared with 57 trajectories (5.7\%) when the
intervention position is selected at random. Among the 500 WebShop and
ScienceWorld trajectories, for which decreases are measurable using a
graded process score, 95 high-disagreement interventions (19.0\%)
degrade trajectory performance.

These results suggest that high-disagreement positions are more likely
to contain critical errors, while local disagreement alone is
insufficient to determine whether a Teacher intervention is beneficial.

\begin{table*}[t]
\centering
\small
\setlength{\tabcolsep}{5.0pt}
\resizebox{\textwidth}{!}{%
\begin{tabular}{lrrrrr}
\toprule
\textbf{Benchmark}
&
\textbf{\# Trajectories}
&
\textbf{High-disagreement Improve}
&
\textbf{No Change}
&
\textbf{Degrade}
&
\textbf{Random Improve}
\\
\midrule
ALFWorld
&
500
&
159 (31.8\%)
&
341 (68.2\%)
&
N/A
&
28 (5.6\%)
\\
WebShop
&
250
&
92 (36.8\%)
&
119 (47.6\%)
&
39 (15.6\%)
&
18 (7.2\%)
\\
ScienceWorld
&
250
&
58 (23.2\%)
&
136 (54.4\%)
&
56 (22.4\%)
&
11 (4.4\%)
\\
\midrule
\multicolumn{6}{l}{
\textit{Aggregated statistics}
}
\\
All benchmarks
&
1,000
&
309 (30.9\%)
&
--
&
--
&
57 (5.7\%)
\\
Graded-score benchmarks
&
500
&
150 (30.0\%)
&
255 (51.0\%)
&
95 (19.0\%)
&
29 (5.8\%)
\\
\bottomrule
\end{tabular}%
}
\caption{
Teacher intervention results on trajectories that fail the terminal
task-success criterion. ALFWorld uses binary terminal success, whereas
WebShop and ScienceWorld additionally use their graded
trajectory-level process scores.
}
\label{tab:motivating_intervention}
\end{table*}

\paragraph{Teacher preference after local intervention.}
We further examine whether a local Teacher intervention changes the
density of positive distillation signals in the subsequent Student
trajectory. Following the Teacher-preferred token ratio defined in the
main paper, we first compute the fraction of Student-generated
continuation tokens for which the Teacher assigns a higher likelihood
than the frozen Student, and then average this ratio across
trajectories.

Before intervention, the mean Teacher-preferred token ratio is
25.45\%. After replacing the Student response at the selected
high-disagreement position with a Teacher response and returning
control to the same frozen Student, the ratio increases to 28.20\%.
This corresponds to an absolute increase of 2.75 percentage points
and a relative increase of 10.8\%, computed from the unrounded values.

This analysis is conducted over all high-disagreement interventions
before applying the future-preference gate. Therefore, the observed
increase is not a direct consequence of selecting trajectories
according to the same Teacher-preference criterion. The result
suggests that local Teacher guidance can shift the subsequent Student
continuation toward regions with denser positive distillation signals.

\paragraph{Training Signal and Evaluation Protocol.}
FTB does not use environment rewards as online optimization signals
or in candidate localization and future-trajectory validation.
Following TCOD-B2F, however, the temporal curriculum is initialized
from pre-collected successful reference trajectories. Thus, beyond
the success-based trajectory selection inherited from B2F, FTB uses
distillation signals derived from the Teacher and Student policies
during interactive training.

Candidate localization uses sampled Teacher--Student disagreement,
and bridge retention is determined by the Teacher-preferred token
ratio along paired Student continuations. Environment outcomes,
including the benchmark-specific signals used in the motivating
intervention analysis above, are used only for downstream performance
evaluation and post-hoc analysis. They are not incorporated as
additional reward signals into the FTB-specific candidate localization,
bridge retention, or optimization procedure.

This protocol allows the comparison with TCOD-B2F to isolate the
effect of future-trajectory validation: FTB introduces no additional
reward-based training supervision beyond the successful reference
trajectories already used by the underlying B2F curriculum.

In FTB, the Teacher-preferred token ratio measures the density of
positive distillation signals along the subsequent Student
continuation. It is used to select Teacher guidance that is consistent
with the OPD objective, and its effectiveness is evaluated through the
downstream performance of the resulting Student policy.

\section{Experimental Details}
\label{sec:appendix_implementation}

This section provides the benchmark splits, model configurations,
generation parameters, optimization settings, and implementation
details used in our experiments. Within each benchmark and
Teacher--Student configuration, all methods use the same Student
initialization, Teacher checkpoint, training steps, and evaluation
setup.

\subsection{Benchmarks and Evaluation}
\label{sec:appendix_benchmarks}

We evaluate all methods on ALFWorld, WebShop, and ScienceWorld. The
data splits, interaction limits, and evaluation metrics are summarized
in Table~\ref{tab:benchmark_settings}.

\begin{table*}[t]
\centering
\small
\setlength{\tabcolsep}{5.0pt}
\resizebox{\textwidth}{!}{%
\begin{tabular}{lllll}
\toprule
\textbf{Benchmark}
&
\textbf{Training Data}
&
\textbf{Evaluation Data}
&
\textbf{Max. Steps}
&
\textbf{Metrics}
\\
\midrule
ALFWorld
&
Standard training split
&
Unseen evaluation split
&
30
&
Success rate
\\

WebShop
&
Subset of the training split (aligned with TCOD)
&
100 held-out sessions
&
15
&
Task score and success rate
\\

ScienceWorld
&
Task-type training split
&
Disjoint task-type split
&
30
&
Task score and success rate
\\
\bottomrule
\end{tabular}%
}
\caption{
Benchmark data, maximum numbers of environment interactions, and
evaluation metrics.
}
\label{tab:benchmark_settings}
\end{table*}

For ALFWorld, success rate is the percentage of episodes completed
successfully. WebShop reports task score and success rate. Let
$r_i\in[0,1]$ denote the environment reward for session $i$. The two
metrics are computed as

\begin{equation}
\mathrm{Score}
=
\frac{100}{N}
\sum_{i=1}^{N} r_i,
\qquad
\mathrm{SR}
=
\frac{100}{N}
\sum_{i=1}^{N}
\mathbf{1}[r_i>0.5].
\end{equation}

The zero-shot and offline WebShop evaluations use 100 sessions with
indices from 4096 to 4195. For ScienceWorld, we report task score and success rate.

\subsection{Models and Training Configuration}
\label{sec:appendix_training}

The main experiments use Qwen3-32B as the Teacher and Qwen3-1.7B or
Qwen3-4B as the Student. All methods within the same experimental
setting share the same frozen Teacher checkpoint and Student
initialization.

The experiments with an RL-trained Teacher use Qwen3-8B as the Teacher
and Qwen3-4B as the Student. The Teacher checkpoint and Student
initialization are fixed across methods in this setting.

\noindent\textbf{Generation and optimization settings.}
All generation processes during training use a temperature of 1.0.
This setting applies to Student rollouts, Teacher prefixes, and Teacher
bridge actions. Student rollouts retain
token-level log-probabilities for OPD supervision, Teacher--Student
disagreement, and continuation alignment. We do not use beam search,
multi-candidate search, or candidate reranking.

Evaluation uses a temperature of 0.4, with thinking mode disabled for
all main results.

We set the distillation coefficient to $\beta=1.0$. Both B2F and
accepted bridge samples are optimized using the same PPO policy loss.
During PPO optimization, the importance ratio is restricted to
$[0.8,1.2]$. For negative advantages, the auxiliary lower-bound factor
is set to 3.0. Losses are averaged over valid response tokens, and the
gradient norm is capped at 1.0.

The shared training hyperparameters are listed in
Table~\ref{tab:training_settings}.

\begin{table}[t]
\centering
\small
\setlength{\tabcolsep}{4.8pt}
\begin{tabular}{lc}
\toprule
\textbf{Hyperparameter} & \textbf{Value} \\
\midrule
Optimizer & AdamW \\
Learning rate & $1\times10^{-6}$ \\
Distillation coefficient $\beta$ & 1.0 \\
Importance-ratio clipping interval & $[0.8,1.2]$ \\
Negative-advantage lower-bound factor & 3.0 \\
Loss aggregation & Token average \\
Rollout batch size & 16 \\
Train batch size & 64 \\
Gradient clipping & 1.0 \\
Training steps & 200 \\
Maximum prompt length & 10,240 tokens \\
Maximum response length & 512 tokens \\
\bottomrule
\end{tabular}
\caption{
Shared generation and optimization settings.
}

\label{tab:training_settings}
\end{table}

\subsection{RL Teacher Training}
\label{sec:appendix_rl_teacher}

We additionally train Qwen3-8B agents with GiGPO to construct stronger
Teachers for the subsequent distillation experiments. Their shared and
benchmark-specific settings are summarized together in
Table~\ref{tab:rl_teacher_settings}.

The maximum numbers of environment interactions are 50 and 15 for
ALFWorld and WebShop, respectively. Validation rollouts use sampling
with a temperature of 0.4, and thinking mode is disabled during RL
training.

\begin{table*}[t]
\centering
\small
\setlength{\tabcolsep}{5.0pt}
\begin{tabular}{lcc}
\toprule
\textbf{Hyperparameter}
&
\textbf{ALFWorld RL Teacher}
&
\textbf{WebShop RL Teacher}
\\
\midrule
Base model
& Qwen3-8B
& Qwen3-8B
\\
RL algorithm
& GiGPO
& GiGPO
\\
Optimizer
& AdamW
& AdamW
\\
Learning rate
& $1\times10^{-6}$
& $1\times10^{-6}$
\\
Weight decay
& 0.01
& 0.01
\\
Discount factor $\gamma$
& 0.95
& 0.95
\\
Rollout group size
& 8
& 8
\\
Invalid-action penalty
& 0.1
& 0.1
\\
Maximum prompt length
& 4,096 tokens
& 4,096 tokens
\\
Maximum response length
& 512 tokens
& 512 tokens
\\
Maximum environment interactions
& 50
& 15
\\
Train batch size
& 32
& 16
\\
PPO mini-batch size
& 64
& 64
\\
PPO micro-batch size per GPU
& 4
& 4
\\
Rollout micro-batch size per GPU
& 4
& 4
\\
Log-prob micro-batch size per GPU
& 8
& 8
\\
Tensor parallel size
& 2
& 2
\\
Number of GPUs
& 8
& 8
\\
\bottomrule
\end{tabular}
\caption{
Configurations used to train the Qwen3-8B RL Teachers for ALFWorld
and WebShop.
}
\label{tab:rl_teacher_settings}
\end{table*}

\subsection{Additional Implementation Details}
\label{sec:appendix_ftb_implementation}

Beyond the procedure described in the main paper, the B2F prefix is
shortened by one environment step every five optimization steps. Each
Student trajectory contains at most one bridge attempt, and the final
Student-controlled turn is excluded because it has no subsequent
continuation for validation.

At the selected position, the Teacher generates one bridge candidate
with a temperature of 1.0. The same frozen Student is then rolled out
for $H=3$ Student-controlled turns on both the original and bridge
branches. Before this paired validation, the environment is
reinitialized and all interactions preceding the selected position are
replayed to recover the same state.

The original B2F samples and accepted bridge samples are optimized
through separate loss terms. Specifically, the overall FTB objective is
\begin{equation}
\mathcal L_{\mathrm{FTB}}
=
\mathcal L_{\mathrm{B2F}}
+
\lambda_{\mathrm{br}}
\mathcal L_{\mathrm{bridge}},
\qquad
\lambda_{\mathrm{br}}=1.
\end{equation}
The B2F loss and bridge loss are independently averaged over their valid
response tokens and then summed with equal coefficients. Bridge
supervision therefore augments rather than replaces the standard B2F
objective, and both sample types use the same optimization settings.

The benchmark-specific prompt templates follow TCOD and are provided
with the released code.

\section{Formal Analysis}
\label{app:formal}

This section complements the definitions in the main paper. We first
clarify the sampled-token reverse-KL estimator, then formalize how the
paired continuation difference estimates the underlying
future-preference advantage, and give a local interpretation of accepted bridge optimization.

\subsection{Sampled-Token Estimation of Reverse KL}

The main paper defines candidate disagreement using tokens sampled by
the Student. For any visited token context $c$,
\begin{equation}
\mathbb{E}_{x\sim\pi_{\bar\theta}(\cdot\mid c)}
\left[
\log
\frac{\pi_{\bar\theta}(x\mid c)}
     {\pi_\phi(x\mid c)}
\right]
=
D_{\mathrm{KL}}
\left(
\pi_{\bar\theta}(\cdot\mid c)
\|
\pi_\phi(\cdot\mid c)
\right).
\label{eq:app_sampled_kl}
\end{equation}
Recent OPD methods commonly estimate the reverse-KL objective through
Student-sampled tokens rather than explicitly summing over the full
vocabulary~\citep{sod,tropd}. FTB follows the same
sampled-token estimator. Its realized value is used only to rank
candidate positions; bridge utility is decided by the paired future
validation.

\subsection{Estimating the Future-Preference Advantage}

Fix the intervention context
\begin{equation}
\mathcal C
=
(h_{t^\star},a^{\mathrm{br}},a^{\mathrm{stu}}).
\end{equation}
Let
\begin{align}
X^{\mathrm{br}}
&=
\rho(\xi^{\mathrm{br}}),
&
X^{\mathrm{base}}
&=
\rho(\xi^{\mathrm{base}}),
\end{align}
where both variables lie in $[0,1]$. The true auxiliary
future-preference advantage of the bridge is
\begin{equation}
\Delta^H_{\phi,\bar\theta}
=
\mathbb E
\left[
X^{\mathrm{br}}-X^{\mathrm{base}}
\mid \mathcal C
\right].
\label{eq:app_true_delta}
\end{equation}
The gate uses the one-pair estimate
\begin{equation}
\widehat\Delta^H_1
=
X^{\mathrm{br}}-X^{\mathrm{base}}.
\label{eq:app_one_pair_delta}
\end{equation}

\paragraph{Proposition 1 (Unbiased paired estimate).}
If the two branches have the intended marginal rollout distributions,
then
\begin{equation}
\mathbb E
\left[
\widehat\Delta^H_1
\mid\mathcal C
\right]
=
\Delta^H_{\phi,\bar\theta}.
\end{equation}

\paragraph{Proof.}
By linearity of conditional expectation,
\begin{align}
\mathbb E
\left[
\widehat\Delta^H_1
\mid\mathcal C
\right]
&=
\mathbb E[X^{\mathrm{br}}\mid\mathcal C]
-
\mathbb E[X^{\mathrm{base}}\mid\mathcal C]
\nonumber\\
&=
\Delta^H_{\phi,\bar\theta}.
\end{align}
\hfill$\square$

The proposition does not require the two branches within a pair to be
independent; it only requires each branch to have the intended
marginal distribution.

To make the relation to the true value more explicit, consider $K$
independent continuation pairs and define
\begin{equation}
\widehat\Delta^H_K
=
\frac{1}{K}
\sum_{k=1}^{K}
\left(
X^{\mathrm{br}}_k-X^{\mathrm{base}}_k
\right).
\end{equation}

\paragraph{Proposition 2 (Concentration around the true advantage).}
For every $\epsilon>0$,
\begin{equation}
\Pr
\left(
\left|
\widehat\Delta^H_K
-
\Delta^H_{\phi,\bar\theta}
\right|
\ge \epsilon
\mid\mathcal C
\right)
\le
2\exp
\left(
-\frac{K\epsilon^2}{2}
\right).
\label{eq:app_delta_concentration}
\end{equation}

\paragraph{Proof.}
For each pair,
$Z_k=X^{\mathrm{br}}_k-X^{\mathrm{base}}_k\in[-1,1]$ and
$\mathbb E[Z_k\mid\mathcal C]=\Delta^H_{\phi,\bar\theta}$.
Applying Hoeffding's inequality to the independent bounded variables
$\{Z_k\}_{k=1}^{K}$ gives
Eq.~\eqref{eq:app_delta_concentration}.
\hfill$\square$

A useful one-sided consequence is that, for any margin $m>0$ and
any fixed context $\mathcal C$ satisfying
$\Delta^H_{\phi,\bar\theta}\le 0$,
\begin{equation}
\Pr
\left(
\widehat\Delta^H_K\ge m
\mid
\mathcal C
\right)
\le
\exp
\left(
-\frac{Km^2}{2}
\right).
\label{eq:app_positive_margin}
\end{equation}
Thus, a larger positive observed bridge gain is exponentially less
likely to arise from a bridge whose true future-preference advantage is
non-positive. With more paired samples, the estimate approaches the
true auxiliary future-preference advantage. FTB uses $K=1$ for efficiency, so the gate remains a noisy estimate.

\subsection{Local Interpretation of Bridge Optimization}

The accepted bridge samples are optimized together with the B2F
samples using the PPO policy loss. Let
\begin{align}
\ell_i(\theta)
&=
\log\pi_\theta
(x_i^{\mathrm{br}}\mid c_i^{\mathrm{br}}),
&
\ell_i^\phi
&=
\log\pi_\phi
(x_i^{\mathrm{br}}\mid c_i^{\mathrm{br}}),
\\
\delta_i(\theta)
&=
\ell_i(\theta)-\ell_i^\phi.
\end{align}
Consider the accepted-token discrepancy
\begin{equation}
\mathcal D_{\mathrm{bridge}}(\theta)
=
\mathbb E
\left[
\frac{\beta g(\tau)}{2M_{\mathrm{br}}}
\sum_{i=1}^{M_{\mathrm{br}}}
\delta_i(\theta)^2
\right].
\label{eq:app_bridge_discrepancy}
\end{equation}

\paragraph{Proposition 3 (First-order bridge update).}
Treat the sampled bridge data, advantages, and gate as fixed. At the
rollout parameters $\theta=\bar\theta$, the gradient of the clipped PPO
loss on accepted bridge tokens equals the gradient of
$\mathcal D_{\mathrm{bridge}}$:
\begin{equation}
\left.
\nabla_\theta
\mathcal L_{\mathrm{PPO}}^{\mathrm{bridge}}(\theta)
\right|_{\bar\theta}
=
\left.
\nabla_\theta
\mathcal D_{\mathrm{bridge}}(\theta)
\right|_{\bar\theta}.
\label{eq:app_bridge_gradient}
\end{equation}

\paragraph{Proof.}
Let
\begin{equation}
r_i(\theta)
=
\frac{
\pi_\theta(x_i^{\mathrm{br}}\mid c_i^{\mathrm{br}})
}{
\pi_{\bar\theta}(x_i^{\mathrm{br}}\mid c_i^{\mathrm{br}})
}.
\end{equation}
At $\theta=\bar\theta$, $r_i=1$ and all ratio-clipping branches agree
locally, while
\begin{equation}
\nabla_\theta r_i(\bar\theta)
=
\nabla_\theta\ell_i(\bar\theta).
\end{equation}
Moreover,
\begin{equation}
A_i^{\mathrm{br}}
=
\beta
\left(
\ell_i^\phi-\ell_i(\bar\theta)
\right)
=
-\beta\delta_i(\bar\theta).
\end{equation}
Therefore, the bridge part of the PPO gradient at the rollout policy is
\begin{align}
\left.
\nabla_\theta
\mathcal L_{\mathrm{PPO}}^{\mathrm{bridge}}(\theta)
\right|_{\bar\theta}
&=
-\mathbb E
\left[
\frac{g(\tau)}{M_{\mathrm{br}}}
\sum_i
A_i^{\mathrm{br}}
\nabla_\theta r_i(\bar\theta)
\right]
\nonumber\\
&=
\mathbb E
\left[
\frac{\beta g(\tau)}{M_{\mathrm{br}}}
\sum_i
\delta_i(\bar\theta)
\nabla_\theta\ell_i(\bar\theta)
\right],
\end{align}
which is exactly
$\nabla_\theta\mathcal D_{\mathrm{bridge}}(\bar\theta)$.
\hfill$\square$

For a sufficiently small gradient step from $\bar\theta$, a first-order
expansion therefore gives
\begin{equation}
\mathcal D_{\mathrm{bridge}}(\theta^+)
=
\mathcal D_{\mathrm{bridge}}(\bar\theta)
-
\eta
\left\|
\nabla_\theta
\mathcal D_{\mathrm{bridge}}(\bar\theta)
\right\|_2^2
+
O(\eta^2).
\end{equation}
Hence, the accepted bridge update locally reduces the sampled
Teacher--Student log-probability discrepancy on the bridge tokens.
This is a local optimization statement and does not imply monotonic
improvement in environment return or global policy divergence.

Taken together, Propositions 1--2 formalize how the observed bridge
gain estimates the true auxiliary future-preference advantage, while
Proposition 3 explains how an accepted bridge locally moves the Student
toward the Teacher on the selected bridge tokens.

\section{Sensitivity to the Validation Horizon}
\label{sec:appendix_horizon_sensitivity}

Beyond the shared training and generation settings, the only
method-specific tunable hyperparameter introduced by FTB is the
validation horizon $H$, which controls the number of
Student-controlled continuation turns used to evaluate a candidate
Teacher bridge. We use $H=3$ in the main experiments and further
evaluate $H\in\{1,3,5\}$ under a fixed 200-step optimization
schedule. These sensitivity results are obtained from single-seed runs
under the main Qwen3-32B Teacher to Qwen3-1.7B Student configuration.

As shown in Table~\ref{tab:horizon_sensitivity}, FTB remains effective
across all tested horizons. The best value of $H$ varies across
benchmarks and evaluation metrics: $H=5$ achieves the highest success
rates on ALFWorld and WebShop, whereas $H=3$ obtains the highest
WebShop score and ScienceWorld success rate. On ScienceWorld, $H=5$
instead achieves the highest task score. The absence of a consistently
dominant horizon, together with the broadly comparable performance
across the tested values, indicates that FTB is not strongly sensitive
to the exact choice of $H$.

\begin{table*}[t]
\centering
\small
\setlength{\tabcolsep}{5.5pt}
\begin{tabular}{c c cc cc}
\toprule
\multirow{2}{*}{\textbf{Horizon $H$}}
&
\textbf{ALFWorld}
&
\multicolumn{2}{c}{\textbf{WebShop}}
&
\multicolumn{2}{c}{\textbf{ScienceWorld}}
\\
\cmidrule(lr){2-2}
\cmidrule(lr){3-4}
\cmidrule(lr){5-6}
&
\textbf{SR $\uparrow$}
&
\textbf{SR $\uparrow$}
&
\textbf{Score $\uparrow$}
&
\textbf{SR $\uparrow$}
&
\textbf{Score $\uparrow$}
\\
\midrule
1
& 39.0
& 53.0
& 55.4
& 3.5
& 27.3
\\
3
& 41.0
& 60.0
& \textbf{60.2}
& \textbf{7.5}
& 26.0
\\
5
& \textbf{43.0}
& \textbf{62.0}
& 56.7
& 4.5
& \textbf{28.9}
\\
\bottomrule
\end{tabular}
\caption{
Single-seed sensitivity to the future-validation horizon $H$ at 200
optimization steps under the main Qwen3-32B Teacher to Qwen3-1.7B
Student configuration. Success rates and task scores are reported on a
0--100 scale.
The best horizon varies across benchmarks and metrics, indicating that
FTB is not strongly sensitive to the exact choice of $H$.
}
\label{tab:horizon_sensitivity}
\end{table*}

\section{Comparison with Closely Related Work}
\label{sec:appendix_concurrent_work}

We focus on methods whose mechanisms are most likely to be confused
with FTB. FTB executes the original Student action and a candidate
Teacher bridge from the same restored environment state. The same
frozen Student then generates both bounded continuations, which are
used only to decide whether the Teacher bridge is retained.

\noindent\textbf{Guided-OPD.}
Guided-OPD mixes Teacher- and Student-generated turns within each
rollout and decreases the Teacher-intervention probability through a
curriculum~\citep{guided-opd}. A selected Teacher turn directly changes
the rollout trajectory. FTB instead preserves the original Student
branch as a comparison and validates each candidate Teacher action
through the Student future it induces.

\noindent\textbf{SAGE-OPD.}
SAGE-OPD uses environment feedback and Teacher judgment to assign
turn-level intervention strengths, and further weights token-level OPD
by Teacher confidence~\citep{SAGE-OPD}. It therefore changes how
strongly an already executed Student response is distilled. FTB
instead generates and executes an alternative Teacher action, then
decides whether that action should be added as auxiliary supervision.

\noindent\textbf{ReOPD.}
ReOPD reuses pre-collected Teacher trajectories as replayed prefixes
and performs Student training without new environment interactions
~\citep{reopd}. Its main design problem is selecting a prefix
distribution that balances Student relevance and Teacher reliability.
FTB operates on states reached during interactive Student rollouts and
uses environment restoration to compare the consequences of two
executable actions.

\noindent\textbf{TOPD.}
TOPD compares short Teacher and Student continuations from the same
textual prefix, uses their near-future divergence to identify genuine
reasoning forks, and injects trajectory discrepancy into a multi-token
objective through trajectory alignment~\citep{topd}. In FTB, both
continuations are generated by the same frozen Student after different
environment actions. The future signal is used as an acceptance gate
rather than as a trajectory-alignment loss.

\noindent\textbf{Relay-OPD.}
Relay-OPD detects a handoff when the Teacher prefers a reflection token
while the Student's top-$K$ support contains no reflection token
~\citep{relayopd}. The Teacher then generates a bounded reasoning leg,
after which the Student may resume, and the resulting relay trajectory
is optimized. FTB does not directly trigger Teacher takeover from this
local asymmetry; it retains a paired original branch and accepts the
Teacher bridge only after comparing the two induced Student futures.

\noindent\textbf{TCOD-B2F.}
TCOD-B2F progressively shortens a successful Teacher or expert prefix
so that the Student takes control from increasingly earlier states
~\citep{tcod}. FTB retains this temporal curriculum and adds
action-level bridge selection, execution, and paired future validation
inside the Student-controlled suffix.

\section{Limitations}

Our experiments use Qwen3 teacher--student pairs with a shared tokenizer, and extending agentic distillation across model families with different tokenizers remains unvalidated. Computational constraints also limit our evaluation to a modest range of model scales, leaving larger teachers and students for future work.

\end{document}